\documentclass[10pt,journal,compsoc]{IEEEtran}
\ifCLASSOPTIONcompsoc
\usepackage[nocompress]{cite}
\else
\usepackage{cite}
\fi

\ifCLASSINFOpdf
\else
\fi
\usepackage{amsmath,amsfonts}
\usepackage{algorithmic}
\usepackage{algorithm}
\usepackage{array}
\usepackage[caption=false,font=normalsize,labelfont=sf,textfont=sf]{subfig}
\usepackage{textcomp}
\usepackage{stfloats}
\usepackage{url}
\usepackage{verbatim}
\usepackage{graphicx}
\usepackage{multirow}
\usepackage{booktabs}
\usepackage{enumitem}
\usepackage{color}
\def\BibTeX{{\rm B\kern-.05em{\sc i\kern-.025em b}\kern-.08em
		T\kern-.1667em\lower.7ex\hbox{E}\kern-.125emX}}

\definecolor{rblue}{rgb}{0,0.5,1}
\definecolor{awesome}{rgb}{1.0, 0.13, 0.32}
\definecolor{hollywoodcerise}{rgb}{0.96, 0.0, 0.63}
\definecolor{lasallegreen}{rgb}{0.03, 0.47, 0.19}
\definecolor{hanpurple}{rgb}{0.32, 0.09, 0.98}
\definecolor{green(pigment)}{rgb}{0.0, 0.65, 0.31}
\usepackage[pagebackref=false,breaklinks=true,colorlinks,bookmarks=false]{hyperref}
\hypersetup{colorlinks=true,linkcolor={red},citecolor={hanpurple},urlcolor={magenta}}

\newcommand{\add}[1]{{\textcolor{black}{#1}}}

\begin{document}

\title{Efficient Point Cloud Processing with High-Dimensional Positional Encoding and Non-Local MLPs}

\author{
	Yanmei~Zou, Hongshan~Yu, Yaonan~Wang, Zhengeng~Yang, Xieyuanli~Chen,\\Kailun~Yang, and Naveed~Akhtar
	\IEEEcompsocitemizethanks{\IEEEcompsocthanksitem Yanmei Zou, Hongshan Yu, Yaonan Wang and Kailun Yang are with the School of Artificial Intelligence and Robotics, Quanzhou Institute of Industrial Design and Machine Intelligence Innovation, Hunan University, Changsha 410082, China. (Yanmei Zou and Hongshan Yu contributed equally to this work. Hongshan Yu is the corresponding author. email: \href{mailto:zouyanmei@hnu.edu.cn}{zouyanmei@hnu.edu.cn},  \href{mailto:yuhongshancn@hotmail.com}{yuhongshancn@hotmail.com}, 	\href{mailto:yaonan@hnu.edu.cn}{yaonan@hnu.edu.cn},
	\href{mailto:kailun.yang@hnu.edu.cn}{kailun.yang@hnu.edu.cn})
	\IEEEcompsocthanksitem Zhengeng Yang is with the College of Engineering and Design, Hunan Normal University, Changsha 410081, China. (email: \href{mailto:yzg050215@163.com}{yzg050215@163.com}).
	\IEEEcompsocthanksitem Xieyuanli Chen is with the College of Intelligence Science and Technology, National University of Defense Technology, Changsha 410073, China.
    \IEEEcompsocthanksitem Naveed Akhtar is with the School of Computing and Information Systems, The University of Melbourne, 3052 Victoria, Australia. (email: \href{mailto:naveed.akhtar1@unimelb.edu.au}{naveed.akhtar1@unimelb.edu.au}).}
}

 \IEEEtitleabstractindextext{ 
\begin{abstract}
Multi-Layer Perceptron (MLP) models are the foundation of contemporary point cloud processing. However, their complex network architectures obscure the source of their strength and limit the application of these models. \add{In this article, we develop a two-stage abstraction and refinement (ABS-REF) view for modular feature extraction in point cloud processing. This view elucidates that whereas the early models focused on ABS stages, the more recent techniques devise sophisticated REF stages to attain performance advantages.} Then, we propose a High-dimensional Positional Encoding (HPE) module to explicitly utilize intrinsic positional information, extending the ``positional encoding'' concept from Transformer literature. HPE can be readily deployed in MLP-based architectures and is compatible with transformer-based methods. Within our ABS-REF view, we rethink local aggregation in MLP-based methods and propose replacing time-consuming local MLP operations, which are used to capture local relationships among neighbors. Instead, we use non-local MLPs for efficient non-local information updates, combined with the proposed HPE for effective local information representation. We leverage our modules to develop HPENets, a suite of MLP networks that follow the ABS-REF paradigm, incorporating a scalable HPE-based REF stage. \add{Extensive experiments on seven public datasets across four different tasks show that HPENets deliver a strong balance between efficiency and effectiveness.} Notably, HPENet surpasses PointNeXt, a strong MLP-based counterpart, by 1.1\% mAcc, 4.0\% mIoU, 1.8\% mIoU, and 0.2\% Cls.~mIoU, with only 50.0\%, 21.5\%, 23.1\%, 44.4\% of FLOPs on ScanObjectNN, S3DIS, ScanNet, and ShapeNetPart, respectively. Source code is available at~\url{https://github.com/zouyanmei/HPENet_v2.git}.
\end{abstract}

}

\maketitle
\IEEEdisplaynontitleabstractindextext
\IEEEpeerreviewmaketitle

\IEEEraisesectionheading{\section{Introduction}\label{sec:introduction}}
\IEEEPARstart{T}{he} increasing popularity of 3D sensors is currently fueling a wide use of 3D point clouds in numerous application domains, such as autonomous driving~\cite{67shi2022weakly}, robotics~\cite{682022Dual} and geological surveying~\cite{101kong2020automatic}. 
Unlike images on regular grids, point clouds consist of irregular 3D points, which pose clear challenges for neural network processing.
%

\begin{figure}[t]
	\centering%
	\includegraphics[width=1.0\linewidth]{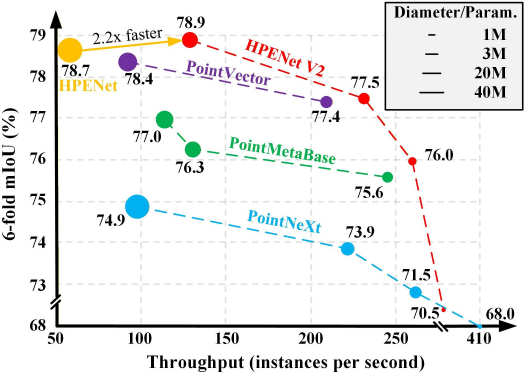}
	\caption{Segmentation performance of HPENet V2 on S3DIS~\cite{43armeni20163d}. 
		\add{Bubble diameter is proportional to model parameter and the legend lists reference sizes of 1M, 3M, 20M, and 40M.}
		HPENet~V2 achieves higher throughput and competitive or better mIoU than state-of-the-art methods, including PointMetaBase~\cite{98lin2023meta}, PointNeXt~\cite{5qian2022pointnext}, and PointVector~\cite{104deng2023pointvector}, while using fewer parameters. 
		Compared with our previous HPENet~\cite{105zou2024improved}, HPENet~V2 delivers comparable performance while using fewer parameters and achieving about $2.2\times$ faster inference.
	}
	\vspace{-3mm}
	\label{fig:fig1}
\end{figure}

Existing neural network based point cloud processing methods can be categorized into two broad categories: voxel-based~\cite{1choy20194d,131peng2024oa} and point-based methods~\cite{3zhao2021point,7qi2017pointnet}. 
Voxel-based methods first discretize 3D space into volumetric units, which improves computational efficiency but inevitably discards fine-grained geometric details.
The seminal work of PointNet~\cite{7qi2017pointnet} originally demonstrated the possibility of directly processing point clouds with MLP-based neural models. Since PointNet, numerous point-based methods have surfaced, e.g., PointNet++~\cite{8qi2017pointnet++}, PointConv~\cite{20wu2019pointconv}, PointNeXt~\cite{5qian2022pointnext}. 
A key attribute of these methods is that they employ sophisticated local aggregation mechanisms to encode locality within point clouds.
For instance, PointNet++ uses a hierarchical network structure for that purpose, whereas PointConv employs a density-aware discrete convolution for high-quality local feature aggregation. The more recent PointNeXt introduces an inverted residual bottleneck module to enhance the scalability of PointNet++.

In recent years, the success of transformers in the natural language processing~\cite{10vaswani2017attention} and computer vision fields~\cite{13liu2021swin} has also motivated transformer-based neural models for directly processing 3D point clouds. To that end, Point Transformer~\cite{3zhao2021point} and other recent methods, e.g., \cite{128li2023ashapeformer,26engel2021point}, use transformer architectures for an even more sophisticated feature aggregation. These efforts are emerging in parallel to the MLP networks for point clouds~\cite{4choe2022pointmixer,18ma2021rethinking}. 
\add{However, these backbones focus on sophisticated, coupled feature extraction pipelines. While PointMetaBase~\cite{98lin2023meta} proposes a general framework (PointMeta), it focuses on local aggregation and ignores sampling operations. In this paper, we show that the key feature extraction modules used by the conventional MLP-based methods and the emerging transformer-based techniques essentially follow the same two-stage abstraction and refinement (ABS-REF) view.} 
As shown in Fig.~\ref{fig:ABS-REF}, the ABS-REF view consists of ABS and REF stages. The ABS stage abstracts lower-resolution features from the input features, and the REF stage refines these abstracted features for better scalability and feature representation without altering the resolution.
Under our ABS-REF view, it becomes clear that whereas the early works, e.g., PointNet++~\cite{8qi2017pointnet++} and PointConv~\cite{20wu2019pointconv}, employ sophisticated local aggregation strategies at the ABS stage, they generally lack the REF stage. Compared with them, the success of the more recent techniques~\cite{3zhao2021point, 4choe2022pointmixer} can be attributed to the REF stage, which enables an increased receptive field of the network and a greater extent of context information considerations. These factors are crucial for discriminative feature learning and thereby improving performance. We discuss this unified view of the latest techniques in Sec.~\ref{sec:PM}. 

Positional information is the key intrinsic geometric property of point clouds, and relative point coordinates can represent the local context between the neighbors and the centroid. However, point-based methods often treat point positions as additional information by concatenating local features with relative point positions, e.g., PointNet++\cite{8qi2017pointnet++} and PointNeXt\cite{5qian2022pointnext}. Though useful, this strategy lacks in giving positional information its due attention. Fortunately, the notion of positional encoding, which originated in the transformer literature~\cite{10vaswani2017attention}, potentially provides an algorithmic solution to this problem by enabling positional information embedding in a feature space. Inspired by this, we propose a High-dimensional Positional Encoding (HPE) for MLP-based point cloud modeling. \add{HPE first projects relative point coordinates into a high-dimensional space, making the encoding translation invariant. This enriched representation is then passed through an MLP to align the high-dimensional vectors to their corresponding feature space. Such high-dimensional projections can improve the performance of coordinate-based MLPs~\cite{148tancik2020fourier}. All processes are packed in an HPE module.} 

Moreover, in MLP-based methods that employ an encoder–decoder architecture, interaction among multi-resolution features is typically unilateral in the decoder, leading to insufficient utilization of contextual information. In this paper, we propose a simple and effective Backward Fusion Module (BFM) to achieve bilateral interaction between multi-resolution features by leveraging the statistical information of high-resolution features. Specifically, BFM first captures contextual information from high-resolution features by using max-pooling and mean-pooling and then embeds captured information into multi-resolution features by inverted residual MLP blocks.

\begin{figure*}[t]
	\centering
	\includegraphics[width=0.9\textwidth]{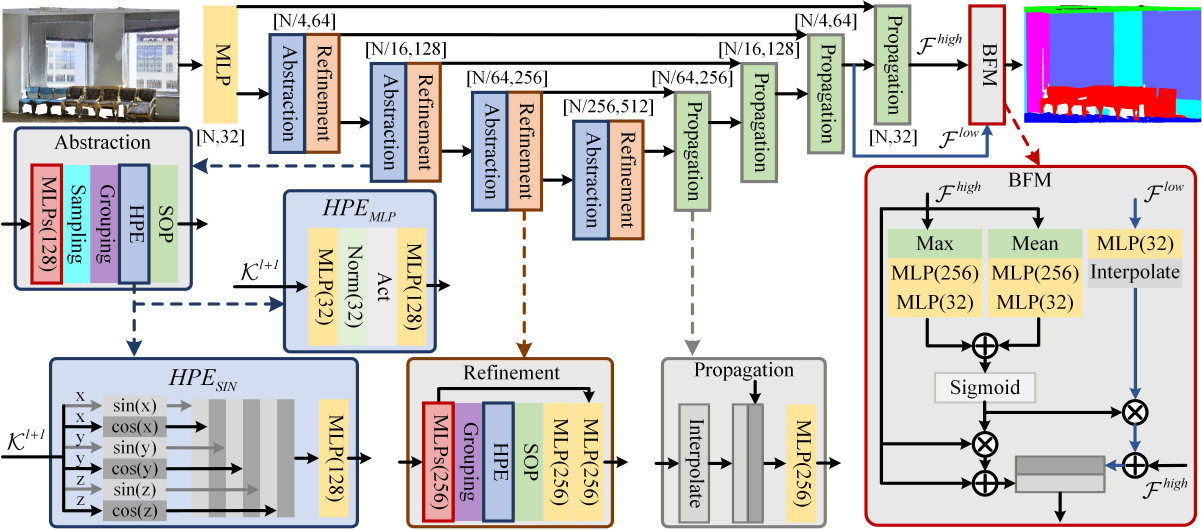}
	\vspace{-1mm}
	\caption{HPENet V2 architecture for semantic segmentation. The network delineates between the Abstraction (ABS) and Refinement (REF) stages of feature extraction and employs the proposed High-dimensional Positional Encoding (HPE) module in both stages. The differences between HPENet V2 and HPENet~\cite{105zou2024improved} are highlighted with \textcolor{red}{red} borders, such as the Backward Fusion Module (BFM) and the non-local MLPs.
	}
	\label{fig:networks}
\end{figure*}

Though the ABS-REF framework can delineate the key strengths of high-performance networks, the complex network architectures and substantial computational requirements render MLP-based methods impractical for real-world applications.
Inspired by efficient depthwise separable convolution from MobileNet~\cite{102howard2017mobilenets}, several MLP-based approaches~\cite{49qian2021assanet,98lin2023meta} decouple local aggregation in intra-set and inter-set operations, which operate on neighboring point sets. Notably, these methods are also consistent with our ABS-REF perspective. 
Building on these efficient methods, we first propose removing traditional, time-consuming local MLPs. 
Then, we reconstruct local aggregation with non-local MLPs, which directly process non-local points before grouping operations, rather than focusing solely on local neighbors as local MLPs do. 
The improved local aggregation significantly reduces FLOPs, but it can compromise performance. 
\add{Hence, following our ABS-REF perspective, we deploy an effective mix of different local aggregation in the ABS and REF stages to leverage the strengths of various MLP types, rather than relying on the commonly adopted monolithic approach~\cite{5qian2022pointnext, 98lin2023meta}.}
The proposed HPE and BFM are used to devise our HPENets, see Fig.~\ref{fig:networks}. 
As shown in Fig.~\ref{fig:fig1}, our method surpasses state-of-the-art (SOTA) methods PointMetaBase~\cite{98lin2023meta},   PointNeXt~\cite{5qian2022pointnext} and  PointVector~\cite{104deng2023pointvector} with parameter reduction and higher throughput.

A preliminary version of this work appeared in the AAAI Conference on Artificial Intelligence~\cite{105zou2024improved}, where we first proposed HPENet, a neural model that employs High-dimensional Positional Encoding (HPE). This article substantially extends~\cite{105zou2024improved} with a focus on efficiency by introducing HPENet V2. As illustrated in Fig.~\ref{fig:fig1}, compared with HPENet, HPENet V2 attains comparable accuracy while markedly reducing computational cost, achieving a 2.2$\times$ speedup and requiring only 0.39$\times$ the parameters. In summary, this work extends~\cite{105zou2024improved} along the following lines.

\begin{itemize}
\item We extend the unified ABS-REF view presented in~\cite{105zou2024improved} for efficient point cloud processing, where non-local MLPs are used to update non-local information, and HPE is deployed to effectively represent local geometric information.  The ABS-REF paradigm provides an intuitive framework for characterizing the key strengths of point cloud processing backbones.
\item 
We propose a simple and effective Backward Fusion Module (BFM) to leverage contextual information. The BFM enables bilateral interaction between multi-resolution features.
\item 
\add{Experiments on MLP-based models are substantially expanded. We rethink various configurations of local aggregation and propose distinct aggregation strategies for different stages, departing from the consistent strategies employed across all stages in prior approaches such as PointNeXt~\cite{5qian2022pointnext} and PointMetaBase~\cite{98lin2023meta}}.
\item
\add{To verify the effectiveness of the proposed modules, we integrate them into two Transformer-based backbones, Point Transformer (PT)~\cite{3zhao2021point} and Stratified Transformer (ST)~\cite{6lai2022stratified}. Our experiments indicate that these modules are highly compatible with both backbones and yield mIoU gains of 2.5\% on PT and 1.3\% on ST.}
\item 
\add{Through extensive evaluation, we achieve SOTA results\footnote{Our claim is limited to the techniques that, similar to our approach, do not benefit from pre-training, voting, or ensembling.} across all tasks studied, including 3D object classification~\cite{42uy2019revisiting,41wu20153d}, scene semantic segmentation~\cite{43armeni20163d,44dai2017scannet, 137behley2019semantickitti}, 3D object part segmentation~\cite{53yi2016scalable}, while also being faster and more device-friendly than existing SOTA MLP-based methods.}
\end{itemize}

\section{Related Work}
Existing point-based methods can be broadly categorized into four groups, namely, MLP-based~\cite{37tolstikhin2021mlp,8qi2017pointnet++,133deng2024ep}, convolution-based~\cite{19engelmann2020dilated,21xu2021paconv}, attention-based~\cite{6lai2022stratified,127he2024full}, and graph-based~\cite{31shen2018mining,118du2024graph,119peigeom,120zhu2021graph,121zhu2020beyond,122lin2021learning,123liang2023long} methods. Key contributions along these categories are discussed below.

\vspace{0.7mm}
\textit{\textbf{MLP-based methods:}} MLP-based methods apply MLPs to extract pointwise features and then use a symmetric operation such as max-pooling or mean-pooling on neighboring point sets to obtain high-level features. After the pioneering work of PointNet~\cite{7qi2017pointnet}, numerous MLP-based techniques have emerged. Most of them focus on devising sophisticated modules to extract the local geometric structure~\cite{8qi2017pointnet++}. Inspired by the widely-used SIFT descriptor~\cite{70lowe2004distinctive}, PointSIFT~\cite{17jiang2018pointsift} develops a 3D SIFT descriptor that considers eight crucial orientations and scales for local scale-invariant feature transform. Building on the simplicity and strong performance of MLP-Mixer~\cite{37tolstikhin2021mlp} in image recognition, PointMixer~\cite{4choe2022pointmixer} adopts a softmax-based approach to effectively mix features within and between point sets, making it highly suitable for the sparse, unordered nature of point clouds.
To improve the generalization and performance of MLP-based networks, PointMLP~\cite{18ma2021rethinking} proposes a local geometric affine module to transform point features in local regions adaptively. 
\add{More recently, ASSANet~\cite{49qian2021assanet}, PointMetaBase~\cite{98lin2023meta}, and DeepLA-Net~\cite{144zeng2025deepla} have proposed computationally efficient solutions for point cloud processing. In particular, DeepLA-Net~\cite{144zeng2025deepla} achieves strong performance by using a compact yet effective local aggregation block within a deep architecture.} Additionally, X-3D~\cite{134sun2024x} explicitly models 3D structure to capture local relationships among neighboring points, \add{and PDNet~\cite{146yin2024point} proposes a deformable point aggregation module for long-range, adaptive information aggregation across points}.

\vspace{0.7mm}
\textit{\textbf{Convolution-based methods:}} These methods focus on designing a local convolution kernel suitable for point cloud processing. For instance, PointConv~\cite{20wu2019pointconv} proposes a density-aware discrete convolution kernel that comprises weight and density functions, whereas KPConv~\cite{2thomas2019kpconv} presents a kernel point convolution that uses any number of kernel points to process various point clouds. 
To overcome the challenges of adapting regular CNNs to irregular point clouds, RS-CNN~\cite{33liu2019relation} introduces a novel convolution method that leverages predefined geometric priors to capture geometric relationships between points, thereby enhancing spatial reasoning and improving shape awareness.
To manage the irregular and unordered nature of point clouds, PAConv~\cite{21xu2021paconv} dynamically constructs convolution kernels by combining basic weight matrices, with combination coefficients adaptively learned from point positions through ScoreNet. 
DPC~\cite{19engelmann2020dilated} employs a dilated point convolution to increase the receptive field size of point convolutional networks. $\Psi$-CNN~\cite{125lei2019octree} proposes a spherical kernel that uses an Octree-guided CNN for the point cloud process. Their method is further enhanced in~\cite{124lei2020spherical} for graph convolution. 

\vspace{0.7mm}
\textit{\textbf{Attention-based methods:}} Attention-based methods exploit attention mechanisms to model long-range dependency between point pairs in a set. These methods are mainly inspired by the attention mechanism~\cite{10vaswani2017attention}, which was first introduced in natural language processing. 
To efficiently process large-scale point clouds, RandLA-Net~\cite{24hu2020randla} uses random point sampling to guarantee efficiency and attention-based local feature aggregation for better performance. Both in natural language processing~\cite{10vaswani2017attention,11devlin2018bert} and computer vision domains~\cite{12dosovitskiy2020image,13liu2021swin}, the attention mechanism is currently causing a paradigm shift. 
Since Point Transformer~\cite{3zhao2021point}, point cloud processing has also started to benefit from this mechanism considerably using the transformer  architectures~\cite{52park2022fast,6lai2022stratified,145zeng2024pointnat,50guo2021pct,114wu2024point}. To address overfitting and limitations on the model depth of Point Transformer~\cite{3zhao2021point}, Point Transformer V2~\cite{90wu2022point} presents grouped vector attention with a grouped weight encoding layer, an additional position encoding multiplier to enhance spatial information embedding, and lightweight partition-based pooling methods for efficient sampling. 
Stratified Transformer~\cite{6lai2022stratified} proposes a stratified key sampling strategy to expand the receptive field while keeping computational costs low and a contextual relative position encoding to adaptively capture spatial information. 
However, transformer-based architectures are computationally intensive, often limiting deployment in real-world applications.

\vspace{0.7mm}
\textit{\textbf{Graph-based methods:}} These methods employ graph structure to extract features, which generally treat points as nodes and feature relations as edges. SPG~\cite{30landrieu2018large} proposes a superpoint graph to deal with large-scale 3D semantic segmentation tasks. DGCNN~\cite{29wang2019dynamic} proposes EdgeConv, which incorporates topology through dynamic graph construction in each network layer to capture local neighborhood information and can be stacked to learn global shape properties. CurveNet~\cite{34xiang2021walk} pays attention to graph structure and employs a shape descriptor, termed ``curves'', using guided walks in point clouds. AGConv~\cite{112wei2023agconv} designs an adaptive kernel in the graph convolution for point cloud processing based on dynamically learned features, making it more flexible than fixed or isotropic kernels and enabling it to effectively capture diverse relationships between points from different semantic parts. 
GSLCN~\cite{123liang2023long} introduces a general graph structure learning architecture to replace the classical approach of using the K-nearest neighbor algorithm for graph construction. 

In summary, point-based methods focus on designing sophisticated modules to effectively capture locality from neighboring point sets.
In contrast, we develop a unified and efficient paradigm with effective geometric representation and a simple and effective feature fusion module.

\section{Proposed Method}
\label{sec:PM}
\begin{figure*}[t]
	\centering
	\includegraphics[width=1\textwidth]{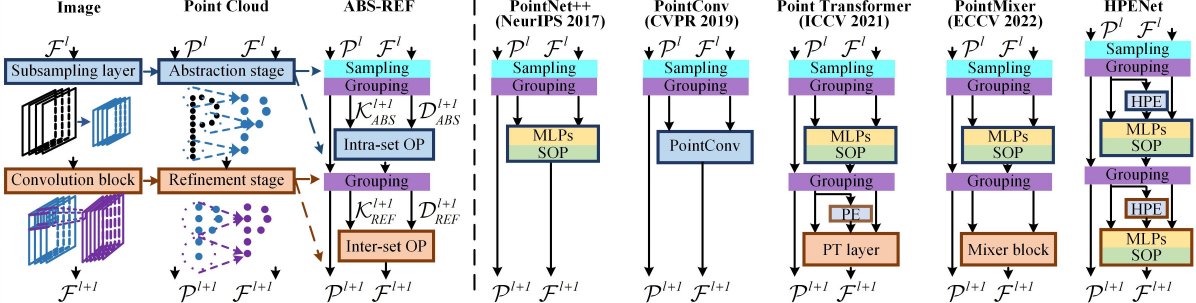}
	\vspace{-6mm}
	\caption{Illustration of abstraction and refinement (ABS-REF) perspective. \textit{Left:} The proposed ABS-REF view of point cloud models is analogous to subsampling and convolution block view in image models. The shown ABS-REF column expands the abstraction and refinement stages. \textit{Right:} Representative instantiations of the ABS-REF framework. Whereas early methods, e.g., PointNet++~\cite{8qi2017pointnet++}, PointConv~\cite{20wu2019pointconv}, ignore the REF stage, more recent techniques, e.g., Point Transformer~\cite{3zhao2021point} and PointMixer~\cite{4choe2022pointmixer}, achieve higher performance by accounting for the REF stage in point cloud models. Abbreviations include SOP: Symmetric OPeration, OP: aggregation OPeration, PT: Point Transformer, and HPE: proposed High-dimensional Positional Encoding.
	}
	\label{fig:ABS-REF}
\end{figure*}


Image processing models are currently experiencing a paradigm shift at the hands of transformers~\cite{12dosovitskiy2020image,13liu2021swin}. Following the suite, many recent works are directly importing the transformer architectures to point cloud  modeling~\cite{3zhao2021point,6lai2022stratified}. However, point cloud data has its peculiar nature. We envisage that a more systematic delineation of the strengths of the existing point cloud techniques can better guide the adoption of relevant concepts of transformers in the point cloud domain. Hence, we first provide a new perspective on the existing point-based methods, then propose a high-dimensional positional encoding enhancement for MLP-based methods, and finally introduce a simple backward attention module to embed contextual information. Below, we first briefly introduce the mathematical notions used in the remaining paper.  

A point cloud with $N$ points can be considered comprising two sets of distinct elements, namely; the point set $\mathcal{P}=\left\{p_m \in \mathbb{R}^{1 \times 3}\right\}_{m=1}^N$ and the feature set $\mathcal{F}=\left\{f_m\in \mathbb{R}^{1 \times C}\right\}_{m=1}^N$, where $p_m$ is the position of the $m$-th point and $f_m$ is the corresponding feature with $C$ channels. In a typical neural model, after a sampling layer, a smaller point cloud is generated with $N^{l+1}$ points, such that  $N^{l+1}< N^l$. Here, $l$ is the index of the sampling layer. By using a grouping operation to group $k$ points neighboring a sampled point in a local region, we get grouped point sets $\mathcal{K}=\left\{k_m \in \mathbb{R}^{k \times 3}\right\}_{m=1}^N$  and the corresponding feature sets $\mathcal{D}=\left\{d_m \in \mathbb{R}^{k \times C}\right\}_{m=1}^N$.

In the following, we first present the details, extension, and instantiation of the unified abstraction and refinement view in Sec.~\ref{abs-ref}. Then, in Sec.~\ref{sec:HPE}, we describe the proposed high-dimensional positional encoding and instantiate it into ${HPE}_{MLP}$ and ${HPE}_{SIN}$ using the dimensional space projection method. Next, in Sec.~\ref{BFM}, we introduce the backward fusion module. Finally, in Sec.~\ref{hpenet}, we present HPENets, incorporating the proposed methods and showing the configuration for point cloud processing. \add{A detailed theoretical analysis of the proposed modules is provided in the supplementary material II and III.}

\begin{figure*}[t]
	\centering
	\includegraphics[width=0.85\textwidth]{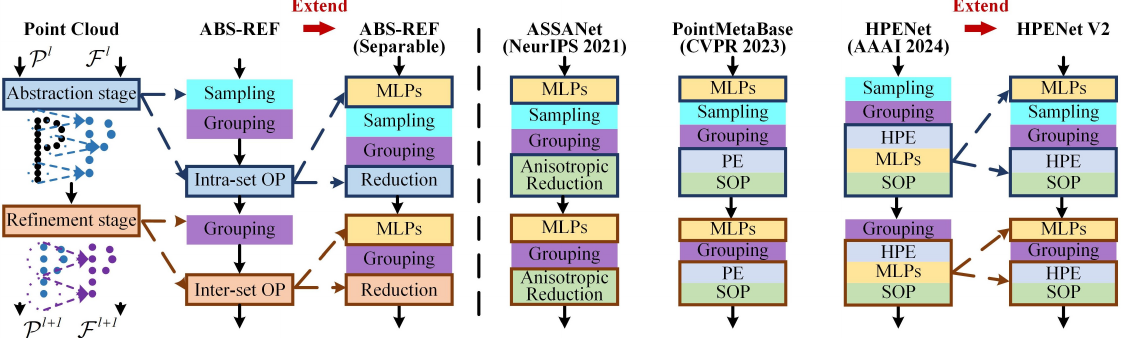}
	\vspace{-1mm}
	\caption{Illustration of separable Abstraction and Refinement (ABS-REF) perspective. \textit{Left:} The separable ABS-REF view also belongs to the proposed ABS-REF by disentangling the intra-set operation and inter-set operation into non-local MLPs and the reduction operation.
		\textit{Right:} Representative instantiations of the separable ABS-REF framework and the evolution of HPENets, i.e., from HPENet to HPENet V2. 
		Please refer to Fig.~\ref{fig:ABS-REF} for the feature propagation in the proposed ABS-REF view.
	}
	\label{fig:ABS-REF-V2}
	\vspace{-3mm}
\end{figure*}

\subsection{Abstraction and Refinement View}
\label{abs-ref}
In Fig.~\ref{fig:ABS-REF} and Fig.~\ref{fig:ABS-REF-V2}, we illustrate the two-stage Abstraction and Refinement (ABS-REF) view of the major existing and the proposed technique. Fig.~\ref{fig:ABS-REF} presents the inspiration behind the ABS-REF view, drawn from image processing. Fig.~\ref{fig:ABS-REF-V2} shows the evolution from a general view to a more balanced view, i.e., the separable ABS-REF view. As previously shown in the image processing domain, traditional feature extractors consist of subsampling and convolutional layers~\cite{135liu2022convnet}. Similarly, we can identify abstraction and refinement layers in point cloud processing methods. \add{However, we find that current point cloud literature generally lacks a clear distinction between the abstraction and refinement processes, which negatively impacts the development of effective and efficient techniques.}

\vspace{0.7mm}
\textit{\textbf{Abstraction (ABS) stage:}}
Analogous to the subsampling operation performed in the image processing networks, we can identify an abstraction (ABS) stage for the point cloud networks. Effectively, this stage eventually abstracts features from the input point cloud and produces a new point cloud with fewer points. The stage can be composed of multiple operations, including a sampling operation (Eq.~\ref{eq:abs sample}), a grouping operation (Eq.~\ref{eq:abs group}), and an intra-set feature aggregation operation (Eq.~\ref{eq:abs aggregation}). Commonly, the sampling operation selects a new point set with fewer elements using Farthest Point Sampling (FPS), which leverages the centroids of local regions for subsampling. The grouping operation generally selects neighboring points around the centroids to define local region sets using, e.g., K-Nearest Neighbors (KNNs). Since the local aggregation operation in the ABS stage abstracts local context information from a set to the corresponding centroid, we call it an intra-set operation. Concretely, given a point set $\mathcal{P}^{l}$ and its corresponding feature set $\mathcal{F}^{l}$, we get the sampled point set $\mathcal{P}^{l+1}$, grouped point sets $\mathcal{K}_{ABS}^{l+1}$, and feature sets $\mathcal{D}_{ABS}^{l+1}$ after sampling and grouping operations. We use the subscript $ABS$ to emphasize the ABStraction stage. In this stage, the intra-set feature aggregation operation $h_{ABS}$ encodes local region patterns into the feature vectors and aggregates intra-set context information. Overall, the abstraction stage can be mathematically expressed as
\begin{equation}
\mathcal{P}^{l+1}=\text{FPS}\left(\mathcal{P}^l\right), p_m^{l+1} \in \mathcal{P}^{l+1},
\label{eq:abs sample}
\end{equation}
\begin{equation}
\mathcal{D}_{ABS}^{l+1}(p_m^{l+1}), \mathcal{K}_{ABS}^{l+1}(p_m^{l+1})=\text{KNN}(p_m^{l+1}, \mathcal{P}^l, \mathcal{F}^l),
\label{eq:abs group}
\end{equation}
\begin{equation}
f_m^{l+1}=h_{ABS}\left(\mathcal{D}_{ABS}^{l+1}\left(p_m^{l+1}\right), \mathcal{K}_{ABS}^{l+1}\left(p_m^{l+1}\right)\right),
\label{eq:abs aggregation}
\end{equation}
where $\mathcal{D}_{ABS}^{l+1}(p_m^{l+1})$ and $\mathcal{K}_{ABS}^{l+1}(p_m^{l+1})$ are the neighbor feature and point sets of the centroid $p_m^{l+1}$, respectively.

Recently, several MLP-based methods~\cite{49qian2021assanet,98lin2023meta} have been proposed to disentangle local aggregation, addressing the need for fast and accurate point cloud processing. As shown in Fig.~\ref{fig:ABS-REF-V2}, these methods, i.e., ASSANet~\cite{49qian2021assanet} and PointMetaBase~\cite{98lin2023meta}, also follow the proposed ABS-REF view. Considering that these methods decouple the Intra-set and Inter-set operations, we refer to these specific instantiations as the separable ABS-REF view. As reported in~\cite{49qian2021assanet} and~\cite{98lin2023meta}, these methods have demonstrated impressive potential for processing point clouds in a mobile-friendly manner, which is also desired in this article.
The separable ABS-REF view disentangles the Intra-set operation into non-local MLPs on channel dimensions and a reduction operation on spatial dimensions, similar to depthwise separable convolution in MobileNet~\cite{102howard2017mobilenets}. Specifically, the non-local MLPs process input points prior to grouping operations, while the reduction operation aggregates local context from neighboring point sets generated by grouping operations.

\vspace{0.7mm}
\textit{\textbf{Refinement (REF) stage:}} Seizing the scalability of convolutional blocks in image processing networks, we can identify a refinement (REF) stage in point cloud networks.
This stage aims to refine the centroid features by aggregating local context information without changing the resolution.
Specifically, the REF stage further processes the point set $\mathcal{P}^{l+1}$ and features set $\mathcal{F}_{ABS}^{l+1}$ generated by the ABS stage. In Fig.~\ref{fig:ABS-REF} (left), we illustrate a simplified architecture of the refinement stage in the adopted ``ABS-REF'' view of the techniques. In the refinement stage, a grouping operation (Eq.~\ref{eq:ref group}) is first used to group the local sets in the centroid point cloud. Later, an inter-set feature aggregation operation $h_{REF}$ is employed to extract and aggregate the inter-set context information. Mathematically, the REF stage can be expressed as
\begin{equation}
\mathcal{D}_{REF}^{l+1}(p_m^{l+1}), \mathcal{K}_{REF}^{l+1}(p_m^{l+1})=\text{KNN}(p_m^{l+1}, \mathcal{P}^{l+1},\mathcal{F}_{ABS}^{l+1}),
\label{eq:ref group}
\end{equation}
\begin{equation}
f_m^{l+1}=h_{REF}\left(\mathcal{D}_{REF}^{l+1}\left(p_m^{l+1}\right), \mathcal{K}_{REF}^{l+1}\left(p_m^{l+1}\right)\right),
\label{eq:ref aggregation}
\end{equation}
where $\mathcal{D}_{REF}^{l+1}(p_m^{l+1})$ and $\mathcal{K}_{REF}^{l+1}(p_m^{l+1})$ are the neighbor feature set and point set of the centroid $p_m^{l+1}$, respectively.

As illustrated in Fig.~\ref{fig:ABS-REF-V2} (left), similar to ABS stages, the separable ABS-REF view also decouples the Inter-set operation in REF stages into non-local MLPs and a reduction operation. Compared to methods in the ABS-REF view, methods within the separable ABS-REF view have lower computational costs~\cite{49qian2021assanet}. For instance, applying MLPs before the grouping operation reduces the FLOPs by  $\frac{C \times C \times N \times k \times L}{C \times C \times N \times L}=k$ times, where $k$ and $L$ are neighborhood size and layer number of MLP, respectively. 

The benefits of joint application of ABS and REF stages in a network are two-fold. 
First, the network gains an effective receptive field from the REF stage. 
Ideally, a centroid's receptive field is $k_{ABS}$ in the ABS stage, while $k_{ABS} \times k_{REF}$ in the REF stage, where $k_{ABS}$ and $k_{REF}$ are the number of neighbor points of a set in the ABS and REF stages. Second, the REF stage helps improve the scalability by increasing the network depth through stacking REF stages, similar to stacking convolutional blocks for images. 

\vspace{0.7mm}
\textit{\textbf{Instantiation of ABS-REF framework:}}
To exemplify a systematic understanding of point cloud models under our ABS-REF perspective, we provide representative examples in Fig.~\ref{fig:ABS-REF} (right) and Fig.~\ref{fig:ABS-REF-V2} (right). It can be seen that early methods like PointNet++~\cite{8qi2017pointnet++} and PointConv~\cite{20wu2019pointconv} only have the ABS stage. Although the two models use different intra-set operations for local feature aggregation, both are single-stage models under our perspective. PointNet++ uses MLPs as intra-set operations, while PointConv utilizes the density-aware discrete convolution. Nevertheless, both models are essentially void of the REF stage. More recently, Point Transformer~\cite{3zhao2021point} and PointMixer~\cite{4choe2022pointmixer} have reported impressive results. Incidentally, we can easily identify an additional REF stage in Point Transformer~\cite{3zhao2021point} and PointMixer~\cite{4choe2022pointmixer}, i.e., positional encoding and Point Transformer layer in~\cite{3zhao2021point} and Mixer block in~\cite{4choe2022pointmixer}. As shown in Fig.~\ref{fig:ABS-REF-V2} (right), the recent ASSANet~\cite{49qian2021assanet} and PointMetaBase~\cite{98lin2023meta} both adhere to the proposed ABS-REF framework, featuring separable intra-set and inter-set operations. Specifically, ASSANet and PointMetaBase decoupled set operations into non-local MLPs and specific reduction operations.

In what follows, we first develop High-dimensional Positional Encoding (HPE) for both ABS and REF stages.
Later, we introduce a simple and effective Backward Fusion Module (BFM) to embed contextual information by fusing multi-resolution features. Thereafter, we leverage HPE and BFM to develop HPENets, which are conveniently designed suites of networks for MLP-based point cloud processing. 
Particularly unique to our models is the inter-set OPeration sub-stage in the REF component, which also distinguishes our technique from the transformer-based methods that employ a REF stage, e.g., Point Transformer~\cite{3zhao2021point}. 

\subsection{High-Dimensional Positional Encoding}
\label{sec:HPE}
Positional information is \textit{the} most important feature of point clouds. It encodes robust geometric details of a scene. Hence, we propose to leverage it fully in both ABS and REF stages of point cloud modeling using explicit Positional Encoding (PE). The notion of PE originated in Transformer literature~\cite{10vaswani2017attention}. 
In the point cloud context, PE can encode a point coordinate $p_m=\left[p_m^x, p_m^y, p_m^z\right] \in \mathbb{R}^{1 \times 3}$  into the space of corresponding feature $f_m \in \mathbb{R}^{1 \times C}$ to embed geometric information. 
For a transformer-based neural architecture for 3D modeling, sinusoidal PE (${PE}_{SIN}$) and learnable PE (${PE}_{MLP}$) can be formulated as below.
\begin{equation}
	PE_{SIN}\left\{\begin{array}{c}
		\left(p_m, 6 i+0\right)=\sin \left(100p_m^x/1000^{6 i/C}\right) \\
		\left(p_m, 6 i+1\right)=\cos \left(100p_m^x/1000^{6 i/C}\right) \\
		\left(p_m, 6 i+2\right)=\sin \left(100p_m^y/1000^{6 i/C}\right), \\
		\left(p_m, 6 i+3\right)=\cos \left(100p_m^y/1000^{6 i/C}\right) \\
		\left(p_m, 6 i+4\right)=\sin \left(100p_m^z/1000^{6 i/C}\right) \\
		\left(p_m, 6 i+5\right)=\cos \left(100p_m^z/1000^{6 i/C}\right)
	\end{array}\right.
\end{equation}
\begin{equation}
	PE_{MLP}\left(p_m\right)=\theta_{3, C}\left(Norm\left(\delta_{3,3}\left(p_m\right)\right)\right).
\end{equation} 
In the above equations, $i=C/6$ is the index of the sub-group PE vector. The $\theta$ and $\delta$ denote MLP-based transformations, with subscripts denoting the channel dimensions of their input and output. $Norm$ denotes the normalization, e.g., batch/layer normalization for restricting the PE to [0, 1]. The sine and cosine functions in ${PE}_{SIN}$ inherently restrict the values in [-1, 1]. Though potentially useful, both ${PE}_{SIN}$ and ${PE}_{MLP}$ provide low-dimensional encodings, which are inadequate to effectively capture the complex geometric relations among the point pairs of unstructured point clouds. Moreover, ${PE}_{SIN}$ is not adaptive. 

To address this limitation, we propose a High-dimensional Positional Encoding (HPE) module. HPE first maps point coordinates into a high-dimensional space to capture fine-grained geometric cues. It then employs a lightweight MLP to align the high-dimensional codes with the feature space, enabling flexible integration into different backbones. \add{The high-dimensional projections can improve the performance of coordinate-based MLPs~\cite{148tancik2020fourier}, and HPE takes relative point coordinates, which are translation invariant. We instantiate HPE with either sinusoidal or learnable encodings, denoted ${HPE}_{SIN}$ and ${HPE}_{MLP}$, respectively.}

Our ${HPE}_{SIN}$ uses sine and cosine functions to extend the channel dimensions from 3 to $(\lfloor C/6\rfloor  \times 6)$ to get a high-dimensional vector, followed by an MLP to align the vector to the feature space. Following the notational conventions from above, ${HPE}_{SIN}$ can be formulated as
\begin{equation}
	HPE_{SIN}\left(p_m\right)=\theta_{(\lfloor C/6\rfloor \times 6), C}\left(PE_{SIN}\left(p_m\right)\right).
\end{equation}

Our $HPE_{MLP}$ generates the high-dimensional vector in a data-driven manner. Specifically, it uses an MLP to extend channel dimensions from 3 to $C$ in HPENet, but to $C/4$ in HPENet V2 for efficiency, followed by another MLP to transform the high-dimensional vector, formulated as
\begin{equation}
	HPE_{MLP}\left(p_m\right)=\theta_{C, C_{mlp}}\left(Norm\left(\delta_{3,C_{mlp}}\left(p_m\right)\right)\right).
\end{equation}

The channel dimension $C_{mlp}$ of the high-dimensional vectors in our encoding can be any suitable value. In HPENet V2, the channel dimension $C_{mlp}$ is reduced to $C/4$ to lower computational costs. Therefore, $HPE_{SIN}$ is not deployed in HPENet V2 since the resulting dimensions $C/4$ are 8 or 16 in the first ABS stage. For a detailed analysis of the effect of $HPE_{SIN}$, please refer to HPENet~\cite{105zou2024improved}.
As shown in Fig.~\ref{fig:networks}, we encapsulate the encoding scheme into $HPE_{SIN}$ and $HPE_{MLP}$ modules, which can be directly plugged into the ABS and REF stages of MLP-based networks. Note that, since there is no sampling operation in the REF stage, the geometric structure of the point cloud remains unchanged. To further reduce computational costs, we share the HPE within each REF stage.

\subsection{Backward Fusion Module}
\label{BFM}
In typical deep neural networks, researchers tend to believe that high-resolution features contain more spatial information, while low-resolution features include more semantic clues~\cite{111qiu2023pointhr}. However, the information interaction of high-resolution feature $\mathcal{F}^{high}$ and low-resolution feature $\mathcal{F}^{low}$ is usually unilateral in the decoder of MLP-based methods, resulting in insufficient use of contextual information. Therefore, in this paper, we introduce a Backward Fusion Module (BFM) to preserve contextual information by reusing low-resolution feature $\mathcal{F}^{low}$ with a bilateral channel attention mechanism for segmentation tasks.

To reduce redundancy from multi-resolution feature fusion, we propose the BFM module, which seizes the design of the classic channel attention mechanism SENet~\cite{93hu2018squeeze}.
As shown in Fig.~\ref{fig:networks}, BFM first employs max-pooling to capture discriminative global features while maintaining a mean-pooling branch, as in SENet, to represent the common patterns of the high-resolution feature $\mathcal{F}^{high}$. Inspired by MobileNet V2~\cite{96sandler2018mobilenetv2}, we utilize MLPs with an expansion layer to refine the statistical contextual information of $\mathcal{F}^{high}$. Subsequently, BFM applies a sigmoid activation, element-wise multiplication, and residual connection to integrate contextual information into the high-resolution feature. Meanwhile, captured global information is propagated backward to the low-resolution feature $\mathcal{F}^{low}$, enabling bilateral interaction between multi-resolution features. Lastly, BFM concatenates updated $\mathcal{F}^{high}$ and $\mathcal{F}^{low}$ for segmentation tasks.

\subsection{HPENets for Point Cloud Processing}
\label{hpenet}
Based on our ABS-REF view and HPE module, we develop MLP-based point cloud processing networks, termed HPENets. To explain, we focus on the more comprehensive encoder-decoder architecture for the semantic segmentation task, as shown in Fig.~\ref{fig:networks}. Other networks can be easily deduced from this explanation (detailed in supplementary material~VI). For object classification, the output features of the encoder are sent to a global pooling layer, followed by a classifier. For object part segmentation, we form a class embedding using the given shape category and concatenate it before the decoder. In Fig.~\ref{fig:networks}, the encoder consists of a single point embedding layer and four blocks that follow the ABS-REF view, while incorporating the proposed HPE modules. The point embedding layer is used to enrich the input representation. We denote the output channels of the point embedding layer as $C_e$, which can be varied. 
The number of REF layers can also vary in the ABS-REF blocks for scalability purposes. 
We denote these numbers by a set $B$ that consists of four elements. To exemplify, our HPENet V2-XL can use $B$ = [3, 6, 3, 3], which means the number of REF layers in the four ABS-REF blocks is 3, 6, 3, and 3, respectively. The value in $B$ can decrease to 0 to degenerate HPENet into a single-stage method, e.g., HPENet V2-S.

As shown in Fig.~\ref{fig:ABS-REF} and Fig.~\ref{fig:ABS-REF-V2}, in the ABS stage, we introduce our HPE module after the grouping layer, which uses the grouped points set $\mathcal{K}_{ABS}^{l+1}$ as the inputs. We first use the grouped feature set $\mathcal{D}_{ABS}^{l+1}$ to add high-dimensional positional encodings and then follow it with a concatenation of grouped points set as the input of local MLPs in HPENet, or directly follow a reduction layer in HPENet V2. 
We opt for a similar strategy in the REF stage. As illustrated in Fig.~\ref{fig:networks}, the obvious difference between the ABS and REF stages is the existence of the sampling layer and the design of local aggregation operation (MLPs). 
In ABS, MLPs are used before the Symmetric OPeration (SOP), i.e., the max pooling operation, as they aim to aggregate local features. 
In contrast, the SOP is embedded between the MLPs in the REF stage. Specifically, the MLP before the SOP pays attention to capturing inter-set context information, while the MLPs following the SOP focus on refining the pointwise features.
By varying the hyper-parameters $B$ and $C_e$, we conveniently construct a range of HPENets with different model sizes to match the training data scales. 
We develop HPENets with the following configurations in our experiments. 
\begin{itemize}
	\item HPENet V2-S: $C_e$ = 32, $B$ = [0, 0, 0, 0].
	\item HPENet V2-B: $C_e$ = 32, $B$ = [1, 2, 1, 1].
	\item HPENet V2-L: $C_e$ = 32, $B$ = [2, 4, 2, 2].
	\item HPENet V2-XL: $C_e$ = 64, $B$ = [3, 6, 3, 3].	
\end{itemize}

\section{Rethinking local aggregation}
\subsection{Local Aggregation in MLP-based Methods}	
MLP-based methods employ local aggregation operators to aggregate local information from neighboring point sets. For simplicity, we focus on analyzing the local aggregation in the ABS stages, which involve more operations. Based on this analysis, the characteristics of local aggregation in the REF stages can be easily deduced.

\begin{figure*}[t]
	\centering
	\includegraphics[width=0.8\textwidth]{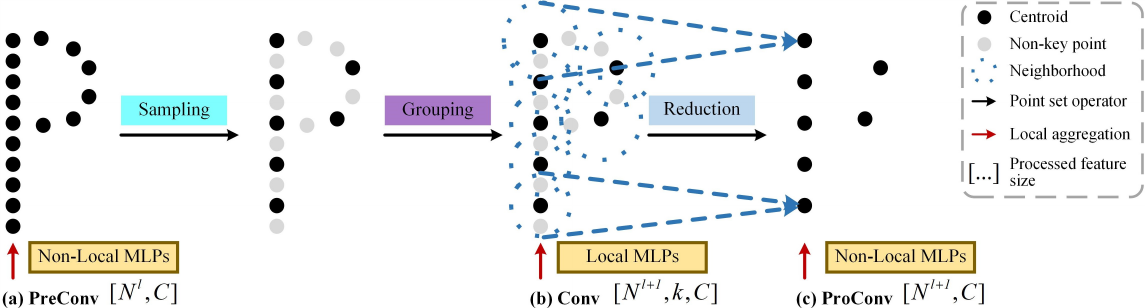}
	\vspace{-2mm}
	\caption{Illustration of local aggregation in the ABS stages of MLP-based methods.
		(a) PreConv processes input point sets with non-local MLPs before sampling operations. (b) Traditional Conv processes neighboring point sets generated by grouping operations with local MLPs. (c) ProConv processes sampled point sets with non-local aggregation after reduction operations.
		Please refer to the text for further notations.
	}
	\label{fig:laconv}
\end{figure*}


In general, local aggregation consists of a grouping operation $G$ to generate neighboring point sets, a set operation $S$ to transform point features, and a reduction layer $R$ to aggregate features from neighboring points. Based on the different point sets processed by set operations, local aggregation operators can be classified into Conv, Preconv, and ProConv.
In Fig.~\ref{fig:laconv}, we illustrate the different local aggregation used in the ABS stages. Conv represents the traditional local aggregation employed in classical MLP-based methods, such as PointNet++~\cite{8qi2017pointnet++} and PointMiXer~\cite{4choe2022pointmixer}, which apply a set operation on neighboring point sets (Eq.~\ref{eq:conv}). PreConv is a balanced local aggregation used in efficient MLP-based methods, such as ASSANet~\cite{49qian2021assanet} and PointMetaBase~\cite{98lin2023meta}, which applies set operations on input point sets (Eq.~\ref{eq:preconv}). Furthermore, ProConv performs set operations on sampled point sets, as defined in Eq.~\ref{eq:preconv}. 
\begin{equation}
	f_m^{l+1}=R(S(G(p_m^{l+1},\mathcal{P}^{l},\mathcal{F}^{l}))),
	\label{eq:conv}
\end{equation}
\begin{equation}
	f_m^{l+1}=R(G(p_m^{l+1},\mathcal{P}^{l},S(\mathcal{F}^{l}))),
	\label{eq:preconv}
\end{equation}
\begin{equation}
	f_m^{l+1}=S(R(G(p_m^{l+1},\mathcal{P}^{l},\mathcal{F}^{l}))),
	\label{eq:proconv}
\end{equation}

Next, we compare local-aggregation variants in terms of the input features processed by MLPs, computational efficiency, and parameter count.

\begin{itemize}
	\item As shown in Fig.~\ref{fig:laconv}, the inputs to PreConv, Conv, and ProConv are, respectively, the input point sets
	$\mathcal{F}^{l}=\{\,f_m\in\mathbb{R}^{C}\,\}_{m=1}^{N^{l}}$, the grouped neighboring point sets
	$\mathcal{D}^{l+1}=\{\,d_m\in\mathbb{R}^{k\times C}\,\}_{m=1}^{N^{l+1}}$, and the sampled point sets
	$\mathcal{F}^{l+1}=\{\,f_m\in\mathbb{R}^{C}\,\}_{m=1}^{N^{l+1}}$.
	The corresponding operating resolutions are $N^{l}$, $N^{l+1}\!\times k$, and $N^{l+1}$.
	Therefore, we term Conv a local MLP design, while PreConv and ProConv are categorized as non-local MLPs.
	PreConv processes the highest-resolution inputs, whereas Conv models local patterns within each neighborhood.
	\item For efficiency, we use FLOPs as the evaluation metric. To simplify and effectively compare the computational costs of different local aggregation operators, we assume that the input and output channels, $C$, and the number of layers, $L$, in the MLPs are consistent across all local aggregation. 
	The FLOPs of PreConv, Conv and ProConv are $C \times C \times N^l \times L$, $C \times C \times N^{l+1} \times k \times L$, and $C \times C \times N^{l+1} \times L$, respectively. Under the proposed ABS-REF framework, sampling operations occur exclusively in the ABS stages, meaning that scale reduction only takes place within the ABS stages. Therefore, the order of FLOPs in the ABS stage is Conv $>$ PreConv $>$ ProConv, while in the REF stage, Conv $= k\times$ PreConv $= k\times$ ProConv.	
	\item For parameter count, all local aggregation operations deploy the same parameters of size $C \times C \times L$, since the channel dimension remains consistent.
\end{itemize}

In summary, PreConv processes the highest-resolution inputs.  Conv captures sophisticated local information that is crucial for point cloud processing, and ProConv provides the most efficient local aggregation design.

\subsection{Experimental Results on Local Aggregation}
In this subsection, we present exploratory experiments on S3DIS Area-5~\cite{43armeni20163d} that examine various local-aggregation strategies under the ABS–REF view, aiming to build an effective and efficient model for point cloud processing.
We present details of the S3DIS dataset and the evaluation protocol in Sec.~\ref{sec:Exp}. Here, we directly discuss relevant results to highlight interesting observations that guide our technique design. 
The relevant results are summarized in Tab.~\ref{tab:conv}, where Conv* only uses neighboring point features as the input of MLPs, whereas Conv deploys the concatenation of neighboring point features and relative point coordinates, similar to PointNet++~\cite{8qi2017pointnet++}. Below, we summarize the interesting observations. 

\vspace{0.7mm}
\textit{\textbf{Firstly}}, Conv is the most effective local aggregation, and positional embedding is crucial for point cloud processing. From lines 1 to 8, it is evident that methods that deploy Conv in both the ABS and REF stages achieve the best performance compared to methods using other configurations. We think there are two main reasons for the significant performance improvement of Conv.
One key factor is the implicit local positional embedding, as Conv outperforms Conv* by 6.1\% in mIoU. 
Another factor is the complex set operations used to capture local information. 
As shown in lines 2, 7, and 8, Conv* outperforms PreConv and ProConv by 1.7\% and 2.6\% in mIoU, respectively.
However, the impact of the positional embedding is more significant than that of the complex set operators. This observation appears to be inconsistent with the common knowledge that many point-based methods focus on developing new and sophisticated modules to extract local structures~\cite{5qian2022pointnext}. Therefore, this finding inspires us to design a more effective positional embedding, i.e., high-dimensional positional encoding.

\vspace{0.7mm}
\textit{\textbf{Secondly}}, considering both performance and efficiency, deploying PreConv in the ABS stage is the most balanced choice. Compared to traditional Conv, PreConv and ProConv improve the efficiency of local aggregation, albeit with some performance reduction. 
As shown in lines 1 and 4, although the method with PreConv reduces the performance of Conv by 1.0\% mIoU, it uses only 73.4\% of the FLOPs required by Conv and is 1.12$\times$ faster. 
Therefore, we consider that this performance degradation of PreConv is acceptable. 
Moreover, a comparison of lines 3 and 4 reveals that PreConv is the more balanced choice, as it outperforms ProConv by 0.6\% in mIoU. Because PreConv processes features at a higher spatial resolution than ProConv, it preserves finer details and captures more contextual information.

\vspace{0.7mm}
\textit{\textbf{Thirdly}}, considering both performance and efficiency, deploying PreConv in the REF stage is the optimal choice. The REF stage enhances scalability by stacking REF stages in a manner similar to stacking convolutional blocks for images. 
Therefore, the resource-intensive Conv is not suitable for deployment in the REF stage when developing efficient networks. As shown in lines 5 and 6, PreConv and ProConv have the same computational costs, whereas PreConv outperforms ProConv by 2.1\% mIoU. 
The main reason is that PreConv can further optimize the features produced by the ABS stage and integrate the refined information into the inter-set operations, while ProConv only updates the output features of the inter-set operations.

Despite its balanced nature, PreConv's performance is impeded by the absence of positional embedding, which is implicitly utilized in Conv. Notably, Positional Encoding (PE), introduced in transformer-based methods~\cite{10vaswani2017attention}, offers a viable solution for effectively incorporating positional information. An analysis of lines 8 and 9 demonstrates that incorporating PE enables PreConv to enhance its performance by 7.9\% mIoU, even surpassing Conv. 
Although PE introduces geometric information into PreConv, its potential is constrained by the limited dimensionality (3 dimensions) of the positional data. We therefore introduce HPE to provide a more effective and robust geometric representation.

Additionally, the local context modeling ability of Conv is significantly stronger than that of PreConv, as information communication in Conv occurs within neighboring point clouds. Meanwhile, the loss of detailed information caused by the sampling operation in the ABS stages is inevitable. To enhance the performance of methods deploying PreConv without substantially increasing computational costs, we propose using Conv only in the first ABS stage to preserve detailed information. 
As shown in line 10 of Tab.~\ref{tab:conv}, our proposed hybrid local aggregation operators achieve a balanced performance with efficiency. These notable performance improvements also illustrate that different local aggregation operators can be employed in the ABS and REF stages, rather than maintaining consistency as previous methods, such as in PointNeXt~\cite{5qian2022pointnext} and PointMetaBase~\cite{98lin2023meta}.
By further introducing the proposed HPE, the performance of our method exceeds the original PreConv baseline by 9.8\% in mIoU, indicating the effectiveness of our design.

\begin{table*}[htbp]
	\centering
	\setlength\tabcolsep{7pt}
	\caption{A study of local aggregation on S3DIS Area-5. \textcolor{black}{$\rm \mathbf{Conv}$, $\rm \mathbf{PreConv}$, and $\rm \mathbf{ProConv}$ represent different local aggregation, $\rm \mathbf{PE}$ refers to Positional Encoding, and $\rm \mathbf{HPE}$ denotes the proposed High-dimensional Positional Encoding. $\mathbf{\Delta}$ represents the difference in mIoU between the current row and the first row. $\rm \mathbf{PreConv^\dagger}$ indicates the use of Conv in the first ABS stage. Please refer to Sec.~\ref{sec:Exp} and Sec.~\ref{sec:sss} for the evaluation metrics.}} %
	\vspace{-1mm}
	\begin{tabular}{@{}cllccccccc@{}}
		\toprule
		Line  & ABS   & REF   & OA (\%)    & mAcc (\%)  & mIoU (\%)   & $\bigtriangleup$ & Param. (M) & FLOPs (G) & TP (ins./sec.) \\
		\midrule
		1    & Conv    & Conv       & 89.0$\pm$0.1  & 73.0$\pm$0.4  & 66.6$\pm$0.4   & -  & 1.58    & 6.93 & 317 \\
		2     & Conv*    & Conv*       &86.3$\pm$0.3  & 67.4$\pm$0.6  & 60.5$\pm$0.6     & -6.1  & 1.58    &  6.76  &  337\\
		3     & PreConv & Conv       & 88.3$\pm$0.4  & 71.9$\pm$0.3 & 65.6$\pm$0.4  & -1.0 &  1.58 & 5.09   & 355  \\
		4     & ProConv & Conv       &  88.1$\pm$0.1  & 71.7$\pm$0.5  & 65.1$\pm$0.4    & -1.5  &  1.58 & 4.90   & 380  \\
		5     & Conv    & PreConv    & 88.6$\pm$0.3  & 72.1$\pm$0.4  & 65.8$\pm$0.6   & -0.8 &1.58   & 3.01 & 388 \\
		6     & Conv    & ProConv    &  87.9$\pm$0.1  & 70.4$\pm$0.4  & 63.7$\pm$0.4  & -2.9 & 1.58   & 3.01 & 388 \\
		7     & ProConv & ProConv    &   85.5$\pm$0.3  & 65.0$\pm$0.8  & 57.9$\pm$0.5     & -8.7 & 1.58   & 0.98   & 492 \\
		8     & PreConv & PreConv    &  85.8$\pm$0.5  & 65.5$\pm$0.3  & 58.8$\pm$0.6  & -7.8  & 1.58   & 1.17   & 452 \\
		9    & PreConv+PE & PreConv+PE & 89.3$\pm$0.3  & 73.4$\pm$0.3  & 66.7$\pm$0.4   & 	\textbf{+0.1} &   1.59 & 1.38   &  328\\
		\midrule
		ours & PreConv$^\dagger$+PE & PreConv+PE & 89.4$\pm$0.2  & 73.6$\pm$0.6  & 67.1$\pm$0.4  & \textbf{+0.5} &1.59  & 1.80 & 324 \\
		ours & PreConv$^\dagger$+HPE & PreConv+HPE &  90.3$\pm$0.1  & 75.0$\pm$0.5  & 68.6$\pm$0.4   & \textbf{+2.0} & 1.76 & 3.47 &  308 \\
		\bottomrule
	\end{tabular}%
	\label{tab:conv}%
\end{table*}%

\section{Experiments}
\label{sec:Exp}
\add{Our technique is extensively evaluated on seven datasets for four different tasks: 3D object classification, scene semantic segmentation, 3D object part segmentation, and 3D object detection.} 
Due to differences in local aggregation and the dimensions of HPE between HPENet V2 and HPENet~\cite{105zou2024improved}, we reconducted experiments to eliminate the impact of structural variations. The ablation study on $HPE_{SIN}$ and $HPE_{MLP}$ is presented in HPENet~\cite{105zou2024improved}. Considering that $HPE_{MLP}$ is more adaptive than $HPE_{SIN}$, we further conduct more detailed ablation studies on $HPE_{MLP}$. Specifically, for efficiency, we set the channel dimension of $HPE_{MLP}$ to $C/4$. \add{For compatibility evaluation, we integrate the proposed modules into the Transformer-based backbones Point Transformer~\cite{3zhao2021point} and Stratified Transformer~\cite{6lai2022stratified}. Details are presented in Sec.~\ref{sec:pt}}.

Unless otherwise specified, we train HPENets using Cross Entropy loss with label smoothing~\cite{45szegedy2016rethinking} and the AdamW optimizer~\cite{46loshchilov2018decoupled} for all tasks. For synthetic object classification on ModelNet40~\cite{41wu20153d}, HPENets are trained with an initial learning rate of 0.001 and a weight decay of 0.05 for 600 epochs. For real-world object classification on ScanObjectNN~\cite{42uy2019revisiting}, following Point-BERT~\cite{47yu2022point}, the number of input points is set to 1,024, with points randomly sampled during training and uniformly sampled during testing. The model is trained with an initial learning rate of 0.002 and a weight decay of 0.05 for 250 epochs.
For indoor semantic segmentation on S3DIS~\cite{43armeni20163d} and ScanNet~\cite{44dai2017scannet}, following Point Transformer~\cite{3zhao2021point}, we evaluate HPENets using the entire scene in each iteration and use the best-performing model for testing. Training is conducted with an initial learning rate of 0.01 for S3DIS and 0.001 for ScanNet, both with a weight decay of 0.0001. \add{For outdoor semantic segmentation on SemanticKITTI~\cite{137behley2019semantickitti}, the learning rate and weight decay for HPENet V2 are set to 0.002 and 0.005, respectively, following Point Transformer V3~\footnote{code: \url{https://github.com/Pointcept/Pointcept.git}}~\cite{114wu2024point}.} For synthetic object part segmentation on ShapeNetPart~\cite{53yi2016scalable}, HPENets are trained with a batch size of 8 for 300 epochs.
For object detection on ScanNet, following VoteNet~\cite{106qi2019deep}, HPENets are trained from scratch for 36 epochs with an initial learning rate of 0.008.

To ensure fair and standard comparisons, we do not conduct any voting~\cite{33liu2019relation} on all datasets except ShapeNetPart. We also compare model parameters (Param.), FLOPs, and inference throughput (TP), measured in instances per second. All baselines are evaluated under a unified protocol. \add{For methods with public implementations, we report the official numbers or recompute them using the released code. For methods without available implementations, the entries are marked with ``–''.}
The TP of all methods is measured using an NVIDIA RTX 4090 24GB GPU and an Intel Xeon Gold 6138 CPU @ 2.00GHz $\times$ 80 cores. The input for TP consists of 128 $\times$ 1024 (i.e., a batch size of 128 and 1024 points) on ScanObjectNN and ModelNet40, 64 $\times$ 2048 on ShapeNetPart, and 16 $\times$ 15000 on S3DIS and ScanNet. \add{Unless otherwise stated, all reported values are means over at least three random seeds. An asterisk (*) denotes the best single-run result obtained under the same training protocol and hyperparameters as the corresponding non-asterisk model.}

\subsection{3D Object Classification}
\textit{\textbf{ScanObjectNN}}~\cite{42uy2019revisiting} collects real-world objects from 700 unique scenes within the SOTA mesh datasets SceneNN~\cite{51hua2016scenenn} and ScanNet~\cite{44dai2017scannet}. This dataset includes about 15,000 real scanned objects, categorized into 15 classes with 2,902 unique object instances. Due to occlusions and noise, ScanObjectNN is a highly challenging dataset for the current methods. Following PointMLP~\cite{18ma2021rethinking}, we evaluate HPENets on PB\_T50\_RS, the hardest and widely used variant of ScanObjectNN, using the standard metrics of mean accuracy (mAcc) and overall accuracy (OA). 

As reported in Tab.~\ref{tab:scanobjectnn}, HPENet V2 significantly improves the efficiency of HPENet with only a slight reduction in performance (a 0.2\% mAcc decrease), while being 1.5$\times$ faster and using only 36.4\% of the FLOPs. Compared with the strong MLP-based method PointNeXt~\cite{5qian2022pointnext}, HPENet V2 surpasses it by 0.7\% in OA and 1.1\% in mAcc, indicating that the proposed HPE can provide effective geometric information and local context for PreConv, which utilizes non-local MLPs. Moreover, HPENet V2 outperforms PointMetaBase~\cite{98lin2023meta}, an MLP-based method that belongs to the separable ABS-REF view, by 0.5\% and 0.7\% in terms of OA and mAcc, respectively. 

\begin{table}[t]
	\centering
	\setlength\tabcolsep{2pt}
	\caption{3D object classification on ScanObjectNN~\cite{42uy2019revisiting}. The best results are highlighted in \textbf{bold}, while the second-best results are \underline{underlined}. Results marked with an asterisk (*) indicate released values.}
	\vspace{-1mm}
	\begin{tabular}{@{}lccccc@{}}
		\toprule
		Methods 					        	  & OA    & mAcc  & Param. & FLOPs & TP\\
		\midrule
		PointNet~\cite{7qi2017pointnet}           & 68.2  & 63.4  & 3.5   & 0.9  & 6846\\
		PointCNN~\cite{16li2018pointcnn}          & 78.5  & 75.1  & 0.6   & - & -\\
		PointNet++~\cite{8qi2017pointnet++}       & 77.9  & 75.4  & 1.5   & 1.7  & 3019\\
		DGCNN~\cite{29wang2019dynamic}            & 78.1  & 73.6  & 1.8   & 4.8  & 1093\\
		PRANet~\cite{57cheng2021net}              & 82.1  & 79.1  & 2.3   & - & -\\
		PointMLP~\cite{18ma2021rethinking}        & 85.7  & 84.4  & 13.2  & 31.4  & 370\\
		PointVector~\cite{104deng2023pointvector} &  87.8$\pm$0.4  & 86.2$\pm$0.5  & 1.6   & \add{1.7} & \add{1484}\\
		\midrule
		PointNeXt-S~\cite{5qian2022pointnext}     & 87.7$\pm$0.4  & 85.8$\pm$0.6  & 1.4   & 1.6  & 4052\\
		PointMetaBase-S~\cite{98lin2023meta}    & 87.9$\pm$0.2   & 86.2$\pm$0.7  & 1.4 & 0.6 & 3140\\
		\midrule
		HPENet~\cite{105zou2024improved}      & \textbf{88.9}  & \textbf{87.6}  &  1.7  & 2.2 & 2700\\
		HPENet (SIN)~\cite{105zou2024improved}      & \underline{88.4}  & 86.9  &  1.7  & 2.2 & 2455 \\
		\midrule
		HPENet V2-S   & 88.4$\pm$0.3      &   86.9$\pm$0.4    & 1.5   & 0.8 & 3989\\
		HPENet V2-S*  &  \textbf{88.9}  &  \underline{87.4}     &    1.5   & 0.8 & 3989\\
		\bottomrule 
	\end{tabular}%
	\label{tab:scanobjectnn}%
\end{table}%

\vspace{0.7mm}
\textit{\textbf{ModelNet40}}~\cite{41wu20153d} is a widely used dataset for synthetic object classification. It contains 12,311 meshed computer-aided design models across 40 different object categories. Following the standard protocol, our training set consists of 9,843 models, while the testing set includes 2,468 models. For evaluation, we utilize both mAcc and OA as metrics.

\begin{table}[t]
	\centering
	\setlength\tabcolsep{3pt}
		
	\caption{3D object classification on ModelNet40~\cite{41wu20153d}.}
	\vspace{-1mm}
	\begin{tabular}{@{}lccccc@{}}
		\toprule
		Methods 							      & OA    & mAcc  & Param. & FLOPs & TP\\
		\midrule
		PointNet~\cite{7qi2017pointnet}           & 89.2  & 86.2  & 3.5   & 0.9   & 6848\\
		PointNet++~\cite{8qi2017pointnet++}       & 91.9  & -     & 1.5   & 1.7   & 3019\\
		PointCNN~\cite{16li2018pointcnn}          & 92.2  & 88.1  & 0.6   &  -   & -\\
		PointConv~\cite{20wu2019pointconv}        & 92.5  & -     & \add{18.6}     & -     & -\\
		DGCNN~\cite{29wang2019dynamic}            & 92.9  & 90.2  & 1.8   & 4.8   & 1093\\
		DeepGCN~\cite{48li2021deepgcns}           & 93.6  & 90.9  & 2.2   & 3.9   & 629\\
		Point Transformer~\cite{3zhao2021point}   & 93.7  & 90.6  & -     & -     & -\\
		PointVector~\cite{104deng2023pointvector} & 93.7  & 91.6  & \add{1.6}    & \add{1.7}     & \add{1486} \\
		CurveNet~\cite{34xiang2021walk}           & 93.8  & 91.1  & \add{2.0}    & -     & -\\
		PointMLP~\cite{18ma2021rethinking}        & \textbf{94.1}  & \underline{91.3}  & 13.2  & 31.4  & 370\\
		\midrule
		PointNeXt-S~\cite{5qian2022pointnext}     & 93.2$\pm$0.1 & 90.8$\pm$0.2  &  1.4  & 1.6   & 4050\\
		PointNeXt-S ($C_e$=64)~\cite{5qian2022pointnext}     &  93.7$\pm$0.3 & 90.9$\pm$0.5 &  4.5 & 6.5   & 1952\\
		\midrule
		HPENet~\cite{105zou2024improved}      &     \underline{94.0}  & 90.8      &  5.9     &  8.7   & 1308 \\
		HPENet (SIN)~\cite{105zou2024improved}      &   \underline{94.0}    & \underline{91.3}      &  5.9     &    8.6  & 1206\\
		\midrule
		HPENet V2-S    &   93.5$\pm$0.2    & 90.8$\pm$0.3      &    1.4   &    0.8   & 3859\\
		HPENet V2-S*    &     93.7  & 90.8   &  1.4     &  0.8     & 3859 \\
		HPENet V2-S ($C_e$=64)    &   93.7$\pm$0.2  & 91.2$\pm$0.4    &    5.1   &    3.1   & 2008\\
		HPENet V2-S ($C_e$=64)*    &  \underline{94.0}     &  \textbf{91.7} & 5.1   &    3.1   & 2008 \\
		\bottomrule 
	\end{tabular}%
	\label{tab:modelnet40}%
\end{table}%

As shown in Tab.~\ref{tab:modelnet40}, HPENet V2 demonstrates consistent improvements over PointNeXt. For instance, HPENet V2-S surpasses PointNeXt-S by 0.3\% in OA. Additionally, HPENet V2 achieves greater resource efficiency with a substantial reduction in FLOPs. Notably, our HPENet V2-S ($C_e$=64) variant achieves SOTA performance, with 94.0\% OA and 91.7\% mAcc on ModelNet40. Moreover, compared to PointMLP~\cite{18ma2021rethinking}, a strong MLP-based method on ModelNet40, HPENet V2-S ($C_e$=64) achieves comparable results with just 5.1M parameters and 3.1G FLOPs, while PointMLP uses 12.6M parameters and 31.4G FLOPs.

\subsection{Scene Semantic Segmentation}
\label{sec:sss}
Scene semantic segmentation aims to assign a semantic label to each point in scene point clouds. This task is generally more challenging than 3D object classification. We evaluate HPENet V2 on two popular large-scale indoor datasets, S3DIS~\cite{43armeni20163d} and ScanNet~\cite{44dai2017scannet}, as well as on the outdoor dataset SemanticKITTI~\cite{137behley2019semantickitti}. The results are summarized in Tables~\ref{tab:s3dis}, \ref{tab:scannet}, and \ref{tab:kitti}, respectively. We discuss these results below.

\begin{table*}[htbp]
	\centering
	\setlength\tabcolsep{9pt}
	\caption{Scene semantic segmentation on S3DIS~\cite{43armeni20163d}.}
	\vspace{-1mm}
	\begin{tabular}{@{}lccccccccl@{}}
		\toprule
		\multirow{2}[4]{*}{Methods} & \multicolumn{3}{c}{S3DIS Area-5} & \multicolumn{3}{c}{S3DIS 6-fold} & Param.    & FLOPs & TP\\
		\cmidrule{2-10}          									& mIoU  & OA    & mAcc  & mIoU  & OA    & mAcc  & M & G & ins./sec. \\
		\midrule
		PointNet++~\cite{8qi2017pointnet++} 		   & 53.5  & 83.0  & -     & 54.5  & 81.0  & 67.1  & 1.0   & 7.2  & 332\\
		PointCNN~\cite{16li2018pointcnn}               & 57.3  & 85.9  & 63.9  & 65.4  & 88.1  & 75.6  & 0.6   & - & -\\
		KPConv~\cite{2thomas2019kpconv}				   & 67.1  & -     & 72.8  & 70.6  & -     & 79.1  & 15.0  & - & -\\
		RepSurf-U~\cite{54ran2022surface} 			   & 68.9  & 90.2  & 76.0  & 74.3  & 89.9  & 82.6  & 1.0   & 1.0  & -\\
		ASSANet~\cite{49qian2021assanet} & 68.0 & 89.7 & 74.3 & - & - & - & 115.6 &  36.2 &- \\
		\add{Fast Point Transformer}~\cite{52park2022fast} 					   & \add{70.1}  & -     & \add{77.4}  & -     & -     & -     &  \add{37.9}     & - & -\\
		BAAF-Net~\cite{58qiu2021semantic} 			   & 65.4  & 88.9  & 73.1  & 72.2  & 88.9  & 83.1  & 5.0   & - & -\\
		CBL~\cite{35tang2022contrastive} &69.4 &90.6 & 75.2 &73.1 &89.6 & 79.4 & 18.6 & -&- \\
		Point Transformer~\cite{3zhao2021point} 	   & 70.4  & 90.8  & 76.5  & 73.5  & 90.2  & 81.9  & 7.8   & 5.6  & -\\
		Stratified Transformer~\cite{6lai2022stratified} & 72.0  & 91.5  & 78.1  & -     & -     & -     & \add{8.0}   &  \add{12.3}  & -\\
		PointVector-L~\cite{104deng2023pointvector}       & 71.2  & 90.8  & 77.3  & 77.4  & 91.4  & 85.5  & 4.2  & 10.7  & \add{209}\\
		PointVector-XL~\cite{104deng2023pointvector}       & 72.3  & 91.0  & 78.1  & 78.4  & 91.9  & 86.1  & 24.1  & 58.5  & \add{90}\\
		Point Transformer V2~\cite{90wu2022point}       & 71.6  & 91.1  & 77.9  & -     & -     & -     & \add{4.4}     & \add{14.1} & - \\
		Point Transformer V3~\cite{114wu2024point} & \textbf{73.4} & \textbf{91.7} &\textbf{78.9} & 77.7 & 91.5 & 85.3 & \add{46.2} & \add{5.2} & - \\
		\midrule
		PointNeXt-S~\cite{5qian2022pointnext}          & 63.4$\pm$0.8 & 87.9$\pm$0.3 & 70.0$\pm$0.7   &   68.0 & 87.4 & 77.3  &0.8   & 3.6   & 410  \\
		PointNeXt-B~\cite{5qian2022pointnext}          &67.3$\pm$0.2 & 89.4$\pm$0.1 & 73.7$\pm$0.6 &  71.5 & 88.8 & 80.2     &3.8   & 8.9   & 293   \\
		PointNeXt-L~\cite{5qian2022pointnext}          &69.0$\pm$0.5 & 90.0$\pm$0.1 & 75.3$\pm$0.8    &  73.9 & 89.8 & 82.2	      & 7.1   & 15.2  & 221\\
		PointNeXt-XL~\cite{5qian2022pointnext}         &70.5$\pm$0.3 & 90.6$\pm$0.2 & 76.8$\pm$0.7 &   74.9 & 90.3 & 83.0  &41.6  & 84.8  & 98   \\
		\midrule
		PointMetaBase-L~\cite{98lin2023meta} 		 & 69.5$\pm$0.3 & 90.5$\pm$0.1  & -     & 75.6 & 90.6  & -     & 2.7   & 2.0   & 245  \\
		PointMetaBase-XL~\cite{98lin2023meta} 		 & 71.1$\pm$0.4 & 90.9$\pm$0.1  & -     & 76.3  & 91.0  & -     & 15.3  & 9.2   & 130   \\
		PointMetaBase-XXL~\cite{98lin2023meta} 		 & 71.3$\pm$0.7 & 90.8$\pm$0.6  & -     & 77.0  & 91.3  & -     & 19.7  & 11.0  & 114  \\
		\midrule
		HPENet~\cite{105zou2024improved}			&    \underline{72.7}   &  {91.5}     &  78.5     &    \underline{78.7}   &  91.9     &  \underline{86.2}   & 46.3  & 148.8& 57 \\
		HPENet (SIN)~\cite{105zou2024improved}			&    72.4   &  91.0     &  \textbf{78.9}    &    78.2   &  \underline{91.7}     &   86.1    &  46.3  & 147.2  &50\\
		\midrule
		HPENet V2-S & 65.2$\pm$0.3   &  89.1$\pm$0.2  & 71.8$\pm$0.4    &   70.5    &  88.7     &    79.6   &  0.7  &  2.6  & 323  \\ 	
		HPENet V2-B & 69.7$\pm$0.3    &  90.4$\pm$0.1 & 	76.0$\pm$0.4    &   76.0 &   91.0    &  84.3     & 1.8  &  3.7  & 260  \\
		HPENet V2-L & 71.0$\pm$0.1  &  91.2$\pm$0.2  & 77.0$\pm$0.3     &  77.5     &  91.6     &  85.3     & 2.9 & 4.2   & 231 \\
		HPENet V2-XL & 72.3$\pm$0.2   &  91.5$\pm$0.1  &  78.4$\pm$0.5  &  78.9     &    91.9   &  86.3     & 16.1  & 18.2  &  128 \\
		HPENet V2-XL* & 72.6 & \underline{91.6}   &  \underline{78.8}   &   \textbf{78.9}     &    \textbf{91.9}   &  \textbf{86.3}    &  16.1  & 18.2  & 128 \\
		\bottomrule 
	\end{tabular}
	\label{tab:s3dis}%
	\vspace{-3mm}
\end{table*}%

\vspace{0.7mm}
\textit{\textbf{S3DIS}}~\cite{43armeni20163d} comprises 6 large-scale indoor areas and 271 rooms, which are captured from 3 different buildings. A total of 273 million points are annotated and classified into 13 semantic categories. Following PointNeXt~\cite{5qian2022pointnext}, we employ two evaluation protocols. The first protocol uses Area-5 as the test scene and all other scenes for training. The second is the standard 6-fold cross-validation. For evaluation, we use the commonly adopted metrics of mean IoU (mIoU), mAcc, and OA. 

From Tab.~\ref{tab:s3dis}, it can be observed that HPENet V2 achieves SOTA performance of 72.6\% mIoU on S3DIS Area-5 and 78.9\% mIoU on S3DIS (6-fold cross-validation). Importantly, no pre-training or voting strategies were used to enhance performance in our results. Detailed semantic segmentation results on S3DIS with 6-fold cross-validation are presented in the supplementary material~IV. Despite being an MLP-based approach, HPENet performs on par with or better than Transformer-based methods. On S3DIS Area-5, HPENet V2-XL surpasses Point Transformer V2~\cite{90wu2022point} by 1.0\% mIoU, 0.5\% OA, and 0.9\% mAcc. \add{Compared with lightweight Transformer-based methods (Fast Point Transformer and Point Transformer V3), HPENet V2 achieves comparable or superior mIoU with substantially fewer parameters (16.1M vs. 37.9M and 46.2M).} Relative to PointNeXt, a strong MLP baseline, HPENet V2 improves mIoU/OA/mAcc by 1.8\%/0.9\%/1.6\% while being 1.3$\times$ faster on Area-5. In the 6-fold setting, HPENet V2 further gains 4.0\% mIoU, 1.6\% OA, and 3.3\% mAcc. Moreover, HPENet V2-L delivers performance comparable to PointMetaBase-XXL while using only 14.7\% of the parameters and 38.2\% of the FLOPs, and it also achieves faster inference.

\begin{figure}[t]
	\begin{center}
		\includegraphics[width=1.0\linewidth]{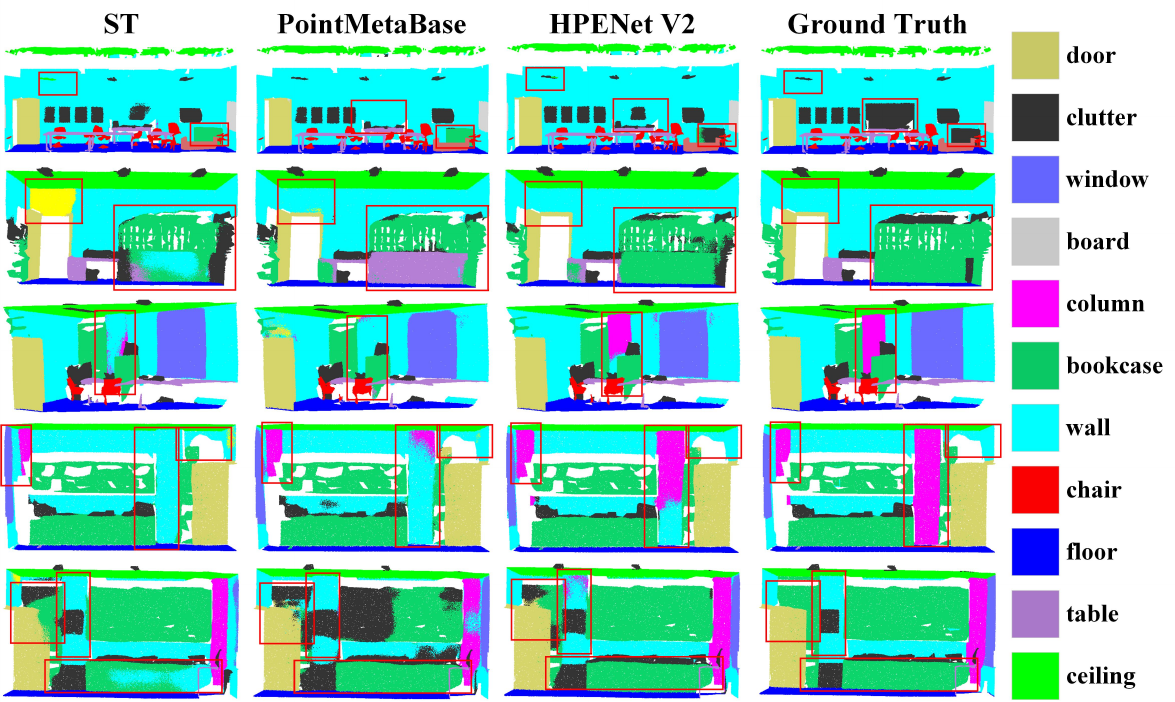}
	\end{center}
	\vspace{-3mm}
	\caption{\add{Representative visual comparison of semantic segmentation results on S3DIS Area-5. “ST” denotes Stratified Transformer.}}
	\label{fig:HPENet}
\end{figure}

\add{Fig.~\ref{fig:HPENet} presents a representative qualitative comparison on S3DIS Area-5 among HPENet~V2, the MLP-based PointMetaBase~\cite{98lin2023meta}, and the transformer-based Stratified Transformer (ST)~\cite{6lai2022stratified}. Relative to these baselines, HPENet~V2 more accurately delineates object categories such as walls ($2^{nd}$ and $5^{th}$ rows), bookcases ($2^{nd}$, $4^{th}$, and $5^{th}$ rows), clutter ($1^{st}$ and $2^{nd}$ rows), and columns ($3^{rd}$ and $4^{th}$ rows). We attribute these improvements to effective modeling of positional information via HPE and fusion of multi-resolution features via BFM.
}

\vspace{0.7mm}
\textit{\textbf{ScanNet V2}}~\cite{44dai2017scannet} consists of 3D indoor scenes with 2.5 million RGB-D frames from over 1,500 scans, annotated with 20 semantic classes. We follow the standard training and validation splits, using 1,201 scenes for training and 312 scenes for validation. The evaluation metric for the validation set is mIoU, following the standard protocol.

As shown in Tab.~\ref{tab:scannet}, HPENet achieves highly competitive performance with 74.0\% mIoU, outperforming PointNeXt by 2.5\% mIoU, though at a higher computational cost. With non-local MLPs, HPENet V2-XL exceeds HPENet (SIN) by 0.5\% mIoU, while using only 21.8\% of the parameters, 9.0\% of the FLOPs, and is 3.6$\times$ faster. Compared to other strong MLP-based methods like PointNeXt and PointMetaBase, our HPENet V2 shows consistent performance gains and is faster. \add{Moreover, compared with the lightweight Transformer-based Fast Point Transformer (FPT)~\cite{52park2022fast}, HPENet~V2 is higher by 0.6\% in mIoU while using less than half the parameters.}

\begin{figure}[htbp]
	\begin{center}
		\includegraphics[width=1.0\linewidth]{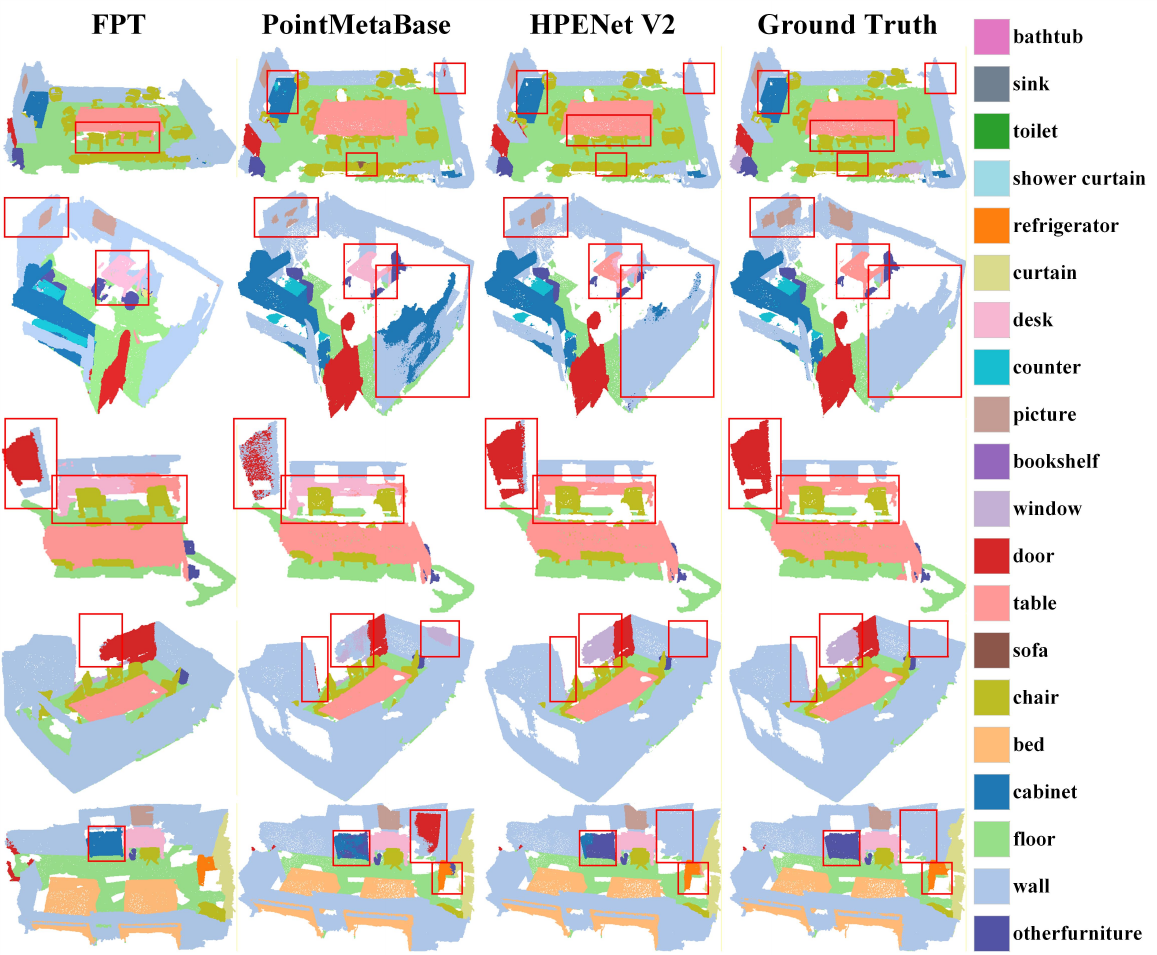}
	\end{center}
	\vspace{-2mm}
	\caption{\add{Representative qualitative comparison of semantic segmentation on ScanNet. Visualizations for Fast Point Transformer (FPT) exclude points annotated with the ignore label.}}
	\label{fig:scannet}
	\vspace{-3mm}
\end{figure}

\add{Fig.~\ref{fig:scannet} presents representative qualitative results on the ScanNet validation set, comparing HPENet V2 with the MLP-based PointMetaBase~\cite{98lin2023meta} and the transformer-based Fast Point Transformer (FPT)~\cite{52park2022fast}. As shown in Fig.~\ref{fig:scannet}, HPENet V2 segments walls ($1^{st}$, $2^{nd}$, $4^{th}$ and $5^{th}$ rows) and tables ($2^{nd}$ and $3^{rd}$ rows) more accurately.}

\begin{table}[t]
	\centering
	\setlength\tabcolsep{6pt}
	\caption{Scene semantic segmentation on ScanNet V2~\cite{44dai2017scannet} val.~set.}
	\begin{tabular}{@{}lcccc@{}}
		\toprule
		Methods & mIoU  & Param. & FLOPs & TP \\
		\midrule
		PointNet++~\cite{8qi2017pointnet++} & 53.5  & 1.0   & 7.2  & 332\\
		RepSurf-U~\cite{54ran2022surface} 	& 70.0  & 1.0   & 1.0  & -\\
		Point Transformer~\cite{3zhao2021point}	& 70.6  & 7.8   & 5.6  &- \\
		MinkowskiNet~\cite{1choy20194d}		& 72.2  & 37.9  & \add{4.7} &- \\
		Fast Point Transformer~\cite{52park2022fast}	& 72.5  & 37.9  & - &- \\
		Stratified Transformer~\cite{6lai2022stratified} & \underline{74.3}  & 18.8  & \add{16.1} & -\\
		Point Transformer V2~\cite{90wu2022point}	& \textbf{75.4}  & 12.8  & \add{18.2} &- \\
		\midrule
		PointNeXt-S~\cite{5qian2022pointnext} 	& 64.5	& 0.7	& 4.7	& 357\\
		PointNeXt-B~\cite{5qian2022pointnext} 	& 68.4	& 3.8	& 9.1	& 289\\
		PointNeXt-L~\cite{5qian2022pointnext} 	& 69.4	& 7.1	& 15.4	& 219\\
		PointNeXt-XL~\cite{5qian2022pointnext} 	& 71.5  & 41.6  & 85.2  & 97\\
		\midrule
		PointMetaBase-L~\cite{98lin2023meta}	& 71.0	& 2.7	& 2.1	& 242\\
		PointMetaBase-XL~\cite{98lin2023meta}	& 71.8	& 15.3	& 9.6 	& 129\\
		PointMetaBase-XXL~\cite{98lin2023meta}	& 72.8 &19.7  & 11.5 & 113\\
		\midrule
		HPENet~\cite{105zou2024improved}	&  74.0 &  74.3  & 220.9 & 40 \\
		HPENet (SIN)~\cite{105zou2024improved}	&  72.8 &  74.2  &  218.6 & 35\\
		\midrule
		HPENet V2-S   &    66.3   &  0.8 & 3.3 &  302\\
		HPENet V2-B   &    70.6   &  1.9 & 4.1 & 255\\
		HPENet V2-L	&   71.4    & 3.0 & 4.6 & 227\\
		HPENet V2-XL	&   73.3    &16.2 & 19.7& 125\\
		\bottomrule
	\end{tabular}%
	\label{tab:scannet}%
\end{table}%

\add{\textit{\textbf{SemanticKITTI}}~\cite{137behley2019semantickitti} is derived from the KITTI Odometry Benchmark and comprises 22 sequences. Sequences 00–10 are used for training, and following common practice~\cite{114wu2024point} we use 08 for validation. Sequences 11–21 are used for testing. Ground-truth labels are released only for the training sequences. The dataset provides dense pointwise annotations for 28 semantic classes on 360$^\circ$ LiDAR scans. Following the official evaluation protocol, we report mIoU over the 19 valid classes.}

\add{As shown in Tab.~\ref{tab:kitti}, HPENet V2-XL achieves 69.3\% mIoU on SemanticKITTI with 18.1G FLOPs. Compared with the transformer-based SphereFormer~\cite{142lai2023spherical}, it gains 1.5\% mIoU while using about $5\times$ fewer FLOPs (18.1G vs. 89.8G) and roughly half the parameters (16.1M vs. 32.3M). These results show that HPENet V2 delivers competitive accuracy at substantially lower computational cost for outdoor semantic segmentation.}

\begin{table}[t]
	\centering
	\setlength\tabcolsep{10pt}
	\caption{\add{Scene semantic segmentation on SemanticKITTI~\cite{137behley2019semantickitti} val.~set.}}
	\begin{tabular}{@{}lccc}
		\toprule
		\add{Methods} & \add{mIoU}  & \add{Param.} & \add{FLOPs}  \\
		\midrule
		\add{SPVNAS}~\cite{138tang2020searching} & \add{64.7} & \add{10.8} & \add{64.5} \\
 		\add{Cylinder3D}~\cite{139zhu2021cylindrical} & \add{64.3} & \add{55.8} & \add{134.2}  \\
 		\add{WaffleIron}~\cite{141puy2023using} & \add{68.0} & - & -\\
 		\add{SphereFormer}~\cite{142lai2023spherical} & \add{67.8} & \add{32.3} &  \add{89.8} \\
 		\add{RangeFormer}~\cite{143kong2023rethinking} & \add{67.6} & \add{24.3} & - \\
 		\add{MinkowskiNet}~\cite{1choy20194d} & \add{63.8} & \add{37.9} & -  \\
 		\add{OA-CNNs}~\cite{131peng2024oa} & \add{70.6} & - & - \\
 		\add{Point Transformer V2}~\cite{90wu2022point} & \add{70.3} & - & -  \\
		\midrule
		\add{HPENet V2-S} & \add{66.9} & \add{0.8} & \add{2.9} \\
		\add{HPENet V2-B} & \add{67.0} & \add{2.2} & \add{4.5}  \\
		\add{HPENet V2-L} & \add{67.4} & \add{3.3} & \add{4.9} \\
		\add{HPENet V2-XL} & \add{69.3} &  \add{16.1} & \add{18.1}  \\
		\bottomrule
	\end{tabular}%
	\label{tab:kitti}%
	\vspace{-3mm}
\end{table}%

\begin{table}[t]
	\centering
	\setlength\tabcolsep{1pt}
	\caption{3D object part segmentation on ShapeNetPart~\cite{53yi2016scalable}.}
	\begin{tabular}{@{}lccccc@{}}
		\toprule
		Methods & Cls. mIoU & Ins. mIoU & Param. & FLOPs & TP\\
		\midrule
		PointNet~\cite{7qi2017pointnet} 					& 80.4  & 83.7  & 3.6   & 4.9  & - \\
		PointNet++~\cite{8qi2017pointnet++} 				& 81.9  & 85.1  & 1.1   & 5.5  & -\\
		DGCNN~\cite{29wang2019dynamic} 						& 82.3  & 85.2  & 1.3   & 12.4  & -\\
		KPConv~\cite{2thomas2019kpconv} 					& 85.1  & 86.4  & -     & - & -\\
		PointMLP~\cite{18ma2021rethinking} 					& 84.6  & 86.1  & -     & - & -\\
		DTNet~\cite{63han2022dual} 						& -     & 86.2  & -     & - & -\\
		AGConv~\cite{112wei2023agconv} & 83.4 & 86.4 & - & - & -\\
		Point Transformer~\cite{3zhao2021point} 			& 83.7  & 86.6  & -     & - & -\\
		Stratified Transformer~\cite{6lai2022stratified} 	& 85.1  & 86.6  & -     & - & -\\
		\midrule
		PointNeXt-S~\cite{5qian2022pointnext} 	         & 84.4$\pm$0.2  & 86.7$\pm$0.0  & 1.0  & 4.5  & 1442\\
		PointNeXt-S (c=64)~\cite{5qian2022pointnext} 	 & 84.8$\pm$0.5  & 86.9$\pm$0.1  & 3.7  & 17.8  & 717\\
		PointNeXt-S (c=160)~\cite{5qian2022pointnext} 	 & 85.2$\pm$0.1  & \underline{87.0$\pm$0.1}  & 22.5  & 110.2  & 225\\
		\midrule
		PointMetaBase-S~\cite{98lin2023meta} 	         & 84.3$\pm$0.1  & 86.7$\pm$0.0  & 1.0 & 1.4  & 1402\\
		PointMetaBase-S(c=64)~\cite{98lin2023meta} 	 &  84.9$\pm$0.2  & 86.9$\pm$0.1  & 3.8 & 3.9  & 797\\
		PointMetaBase-S(c=160)~\cite{98lin2023meta} 	 & 85.1$\pm$0.3  & \underline{87.1$\pm$0.0}  & 22.7 & 18.5  & 314\\
		\midrule
		HPENet~\cite{105zou2024improved}				& 85.3     &  87.0     &   24.7     &  135.6 & 171 \\
		HPENet (SIN)~\cite{105zou2024improved}				& \textbf{85.5}  & \underline{87.1}  &    24.7   &  135.3 & 158\\
		\midrule
		HPENet V2-S 										&   84.6$\pm$0.1    &  86.8$\pm$0.1 &    1.1   & 2.1 & 1567\\
		HPENet V2-S(c=64) 										&  84.8$\pm$0.4     & 86.8$\pm$0.0  &  4.3     & 8.0 & 823\\
		HPENet V2-S(c=160) 										&  85.4$\pm$0.1     &  87.2$\pm$0.1 &    26.3   & 48.9 & 293\\
		HPENet V2-S* 										&  \underline{85.4}     &  \textbf{87.3}     &  26.3   & 48.9 & 293 \\
		\bottomrule
	\end{tabular}%
	\label{tab:shapenetpart}%
\end{table}%

\begin{table*}[t]
	\centering
	\setlength\tabcolsep{8pt}
	\caption{3D object detection results on SUN RGB-D validation set with mAP@0.25. ${}^{\ddagger}$ donates that the model is implemented on MMDetection3D~\cite{r39}.}
	
	\begin{tabular}{@{}lccccccccccc@{}}
		\toprule
		Model          & mAP@0.25     & bed  & table & sofa & chair & toilet & desk & dresser & nightstand & shelf & bathtub    \\ 
		\midrule
		VoteNet~\cite{106qi2019deep}   & 57.7    & 83.0 & 47.3  & 64.0 & 75.3  & 90.1   & 22.0 & 29.8    & 62.2 & 28.8  & 74.4          \\
		VoteNet${}^{\ddagger}$~\cite{106qi2019deep}    &  59.7  &  84.8    &  49.6 & 67.8 & 77.6 &  87.4  & 24.3  & 29.3   & 61.9 &  32.1 &  82.1 \\
		BRNet~\cite{110cheng2021back}   & \underline{61.1}     & 86.9 & 51.8  & 66.4 & 77.4  & 91.3   & 29.6 & 35.9    & 65.9       & 29.7  & 76.2          \\
		3DETR~\cite{108misra2021end}    & 59.1   & 84.6 & 52.6  & 65.3 & 72.4  & 91.0   & 34.3 & 29.6    & 61.4     & 28.5  & 69.8           \\
		\midrule
		VoteNet${}^{\ddagger}$ + $HPE_{MLP}$    &   \textbf{61.2} &   85.0   &  50.5    &   67.7   & 78.5     & 91.1 &  28.5 & 35.0 & 64.1 & 34.3 &  77.1  \\ 
		\bottomrule
	\end{tabular}%
	\label{tab:detection2}%

\end{table*}%

\subsection{3D Object Part Segmentation}
ShapeNetPart~\cite{53yi2016scalable} is an object-level dataset for object part segmentation, consisting of 16,881 objects from 16 shape categories and 50 part labels. Following PointNet++~\cite{8qi2017pointnet++}, we randomly select 2,048 points as input and evaluate performance using class mean IoU (Cls. mIoU) and instance mean IoU (Ins. mIoU).

In Tab.~\ref{tab:shapenetpart}, we report the results of the top-performing approaches. HPENet V2 outperforms the SOTA method and is also faster on this dataset. Our HPENet V2 reaches 87.3\% Ins. mIoU, while the performance of MLP-based methods has remained stagnant below this value for years. Notably, HPENet outperforms the strong transformer-based method Stratified Transformer~\cite{6lai2022stratified} not only on S3DIS but also on ShapeNetPart.

\begin{figure}[t]
	\begin{center}
		\includegraphics[width=0.9\linewidth]{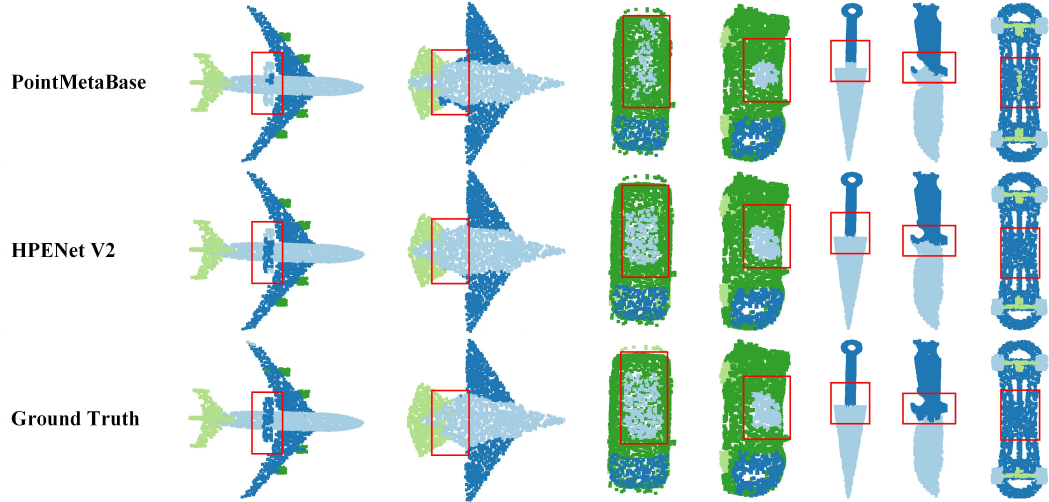}
		\vspace{-2mm}
	\end{center}
	\caption{Qualitative comparisons among PointMetaBase~\cite{98lin2023meta} (first row), HPENet (second row), and Ground Truth (third row) on ShapeNetPart~\cite{53yi2016scalable} for 3D object part segmentation.}
	\label{fig:shapenetpart}
\end{figure}

Fig.~\ref{fig:shapenetpart} presents qualitative results of PoinMetaBase~\cite{98lin2023meta} and HPENet V2 on ShapeNetPart. In Fig.~\ref{fig:shapenetpart}, HPENet V2 produces
predictions that are closer to the ground truth compared to PointMetaBase. Specifically, HPENet V2 accurately segments the wings ($1^{st}$ column), bodies and engines ($2^{nd}$ column) of airplanes, roofs ($3^{rd}$ and $4^{th}$ columns) of cars, handles ($5^{th}$ and $6^{th}$ columns) of knifes, and bearing ($7^{th}$ column) of a skateboard.

\subsection{3D Object Detection}
The key building block of HPENets, i.e., the HPE module, is inherently compatible with MLP-based backbones. 
To examine its flexibility, we also extend the competitive techniques of VoteNet~\cite{106qi2019deep} and GroupFree3D~\cite{107liu2021group} with $HPE_{MLP}$. 
In Tab.~\ref{tab:detection2} and Tab.~\ref{tab:detection1}, we summarize the results of our extension following standard evaluation protocols on ScanNet V2~\cite{44dai2017scannet} and SUN-RGB-D~\cite{109song2015sun} for 3D object detection. 

ScanNet V2 contains richly annotated 3D reconstructed indoor scenes with axis-aligned bounding boxes for object detection. It includes 1,201 training samples and 312 scans for validation. SUN-RGB-D is a single-view indoor dataset consisting of 10,355 RGB-D images and contains 5,000 training images annotated with oriented 3D bounding boxes. As shown in Tab.~\ref{tab:detection2} and Tab.~\ref{tab:detection1}, a consistent gain across the board is achieved with our $HPE_{MLP}$ extension.

\begin{table*}[htbp]
	\centering
	\setlength\tabcolsep{14pt}
	\caption{Ablation study on S3DIS Area-5 demonstrating the efficacy of ABS-REF view, and contribution of HPE modules. $\mathbf{\Delta}$ represents the difference in mIoU between the current row and the first row.}
	\begin{tabular}{@{}lcccccccc@{}}
		\toprule
		Networks($C_e$, $B$) & HPE  & OA & mAcc & mIoU  &  $\triangle$ & Param. & FLOPs & TP\\
		\midrule
		HPENet\_dv (32, [0, 0, 0, 0]) 	&         & 82.7$\pm$0.6  & 58.3$\pm$0.7  & 51.5$\pm$0.2  & - & 0.5 & 0.8 & 509 \\

		HPENet\_dv (32, [1, 1, 1, 1]) 	&         &  85.8$\pm$0.5 & 65.5$\pm$0.3& 58.8$\pm$0.6 & +7.3 & 1.6 & 1.2 & 452  \\
		HPENet\_dv (32, [1, 1, 1, 1]) 	& PE &89.3$\pm$0.3 & 73.4$\pm$0.3  & 66.7$\pm$0.4 & +15.2 & 1.6 & 1.4   &  328 \\
		HPENet\_dv (32, [1, 1, 1, 1]) 	& HPE & 90.0$\pm$0.1  & 74.5$\pm$0.8  & 68.1$\pm$0.4  & +16.6 & 1.8 & 3.2 & 309 \\
		\midrule
		HPENet\_dv${}^{\star}$					&         & 83.5$\pm$0.5  & 60.1$\pm$0.8  & 53.2$\pm$0.7  & -5.6  & 1.6 & 1.2 & 507 \\
		\bottomrule 
	\end{tabular}%
	\label{tab:abs-ref}%
	\vspace{-2mm}
\end{table*}%

\begin{table}[t]
	\centering
	\setlength\tabcolsep{8pt}
	\caption{3D object detection on ScanNet V2 validation
		set. ${}^{\ddagger}$ donates that the model is implemented on MMDetection3D~\cite{r39}.}
	\begin{tabular}{@{}lcc@{}}
		\toprule
		Method  & mAP@0.25 & mAP@0.5 \\
		\midrule
		VoteNet~\cite{106qi2019deep}       & 58.6     & 33.5    \\ 
		VoteNet${}^{\ddagger}$~\cite{106qi2019deep}       & 63.8     & 44.2    \\
		MLCVNet~\cite{136xie2020mlcvnet}       & 64.5     & 41.4    \\ 
		3DETR~\cite{108misra2021end}       & 65.0    & 47.0    \\
		GroupFree3D${}^{\ddagger}$~\cite{107liu2021group}   & 68.2    & \underline{52.6}    \\ 
		GroupFree3D + X-3D~\cite{134sun2024x} & \underline{69.0} & 51.1 \\
		\midrule
		VoteNet + $HPE_{MLP}$ & 60.9    & 37.4  \\
		VoteNet${}^{\ddagger}$ + $HPE_{MLP}$ & 65.0     & 45.6    \\
		GroupFree3D${}^{\ddagger}$ + $HPE_{MLP}$ & \textbf{69.1}     & \textbf{53.0} \\
		\bottomrule
	\end{tabular}%
	\label{tab:detection1}%
	\vspace{-3mm}
\end{table}%

\begin{table*}[htbp]
	\centering
	\setlength\tabcolsep{8pt}
	\caption{Ablation study for positional encoding and backward fusion module with HPENet V2-XL on S3DIS Area-5 justifying HPE use in both ABS and REF stages. $\mathbf{\Delta}$ in lines 1 through 5 indicates the difference in mIoU between the corresponding row and the first row, while $\mathbf{\Delta}$ in lines 6 through 8 represents the difference in mIoU between the corresponding row and the fourth row.}
	\begin{tabular}{@{}clcccccccccc@{}}
		\toprule
		Line & Type  & ABS   & REF   & BFM    & OA    & mAcc & mIoU  & $\triangle$ & Param.(M) & FLOPs(G) & TP(ins./sec.) \\
		\midrule
		1&  & & & &87.5$\pm$0.4  & 70.4$\pm$0.5  & 63.9$\pm$0.3   & - &15.4 & 10.4& 176\\
		2& HPE	  & $\surd$ &         &          &  90.2$\pm$0.1  & 76.1$\pm$0.4  & 70.0$\pm$0.4      &    +6.1   &    15.8   &  13.7     & 160 \\
		3& HPE	  &         & $\surd$ &          &   91.0$\pm$0.1  & 76.9$\pm$0.6  & 70.7$\pm$0.5    &   +6.8    &      15.8 &  14.5     & 148 \\
		4&	HPE  & $\surd$ & $\surd$ &  & 91.5$\pm$0.2  & 77.9$\pm$0.3  & 72.2$\pm$0.2   &    +8.3     &    16.0     &   17.7     & 137  \\
		5&	HPE  & $\surd$ & $\surd$ & $\surd$  & 91.5$\pm$0.1   & 78.4$\pm$0.5  &   72.3$\pm$0.2&  +8.4  & 16.1  &  18.2  &  128  \\
		\midrule
		6&PE    							  & $\surd$ & $\surd$ &          &  90.9$\pm$0.2  & 77.0$\pm$0.5  & 70.9$\pm$0.5   &    -1.3   &   15.4     &   10.8    &    144 \\
		7&HPE (mul)				  & $\surd$ & $\surd$ &          &     90.9$\pm$0.1  & 76.8$\pm$0.2  & 71.0$\pm$0.5   & -1.2   & 16.0     &   17.7     & 137  \\
		8&HPE (abs) 			  & $\surd$ & $\surd$ &          &     78.3$\pm$1.8  & 63.2$\pm$1.0  & 54.6$\pm$1.6      & -17.6  &  16.0    &     10.7  &  175 \\
		\bottomrule 
	\end{tabular}%
	\label{tab:hpe}%
\end{table*}%

\begin{table}[htbp]
	\centering
	\setlength\tabcolsep{3pt}
	\caption{Ablation study on dimension of $HPE_{MLP}$ with HPENet V2-XL on S3DIS Area-5. `C' denotes the feature channel number. $\mathbf{\Delta}$ represents the difference in mIoU between the current row and the first row.}
	\begin{tabular}{@{}lccccccc@{}}
		\toprule
		Dimension   & OA    & mAcc  & mIoU & $\triangle$ & Param. & FLOPs & TP \\
		\midrule
		3     	  & 90.9$\pm$0.2  & 77.0$\pm$0.5  & 70.9$\pm$0.5    & - & 15.3     &   10.8    &    144   \\
		C/8      &   91.6$\pm$0.2  & 77.9$\pm$0.2  & 71.9$\pm$0.3      &   +1.0    &     15.6  &  14.1     & 141 \\
		C/4      &  91.5$\pm$0.2  & 77.9$\pm$0.3  & 72.2$\pm$0.2   &     +1.3   &    16.0     &   17.7     & 137 \\
		C/2      &   91.4$\pm$0.1  & 77.4$\pm$0.2  & 71.5$\pm$0.1  &    +0.6   &   16.7    & 24.9      &  129\\
		C         & 91.4$\pm$0.3  & 78.2$\pm$0.7  & 71.8$\pm$0.4      &   +0.9    & 18.1      &  39.3  & 116 \\
		\bottomrule 
	\end{tabular}%
	\label{tab:hpe-c}%
	\vspace{-3mm}
\end{table}%

\begin{table*}[htbp]
	\centering
	\setlength\tabcolsep{5pt}
	\caption{\add{Ablation of the proposed modules on Point Transformer (PT)~\cite{3zhao2021point} and Stratified Transformer (ST)~\cite{6lai2022stratified} for S3DIS Area-5. $\Delta$ denotes the mIoU difference relative to the baseline row. cRPE (Contextual Relative Position Encoding) is a \emph{spatial} relative positional encoding used in ST.}}
	\begin{tabular}{@{}llcccccccc@{}}
		\toprule
		Methods &  ABS (MLPs)   & REF (PT layers) & BFM & OA & mACC  & mIoU  &  $\triangle$  & Param.(M) & FLOPs(G) \\
		\midrule
		Point Transformer~\cite{3zhao2021point} &  Local MLPs   & PE & & 90.4 & 74.8  & 68.2   & - &   7.8    &  5.6     \\
		+Non-local MLPs & Non-local MLPs  &  PE & &  90.7  & 76.3  & 69.7    & +1.5 &   7.8     &    4.9    \\
		+HPE  & Non-local MLPs + HPE & HPE&  & 90.5  &  76.4  &  70.3 & +2.1 &     7.8 &     5.2 \\
		+BFM & Non-local MLPs + HPE & HPE & $\surd$ &   90.9  & 76.6  & 70.7   & +2.5 &  7.9     &    5.3  \\
		\midrule
		\add{Stratified Transformer}~\cite{6lai2022stratified} &  \add{Local MLPs}   & \add{cRPE} &  & \add{91.7}  & \add{77.4}  & \add{71.4}     &  -  & \add{8.0}   &   \add{12.3}  \\
		\add{+Non-local MLPs} & \add{Non-local MLPs} &  \add{cRPE} & & \add{90.9}  & \add{78.9}   & \add{72.0}  &          \add{+0.6}   & \add{8.0}    & \add{11.1}    \\
		\add{+HPE}  & \add{Non-local MLPs + HPE} & \add{cRPE} &    &  \add{92.2} & \add{78.7}   & \add{72.5}    & \add{+1.1}  &       \add{8.1}  & \add{11.9}       \\
		\add{+BFM} & \add{Non-local MLPs + HPE} & \add{cRPE} & \add{$\surd$} & \add{91.9}   & \add{78.7}  & \add{72.7}   &  \add{+1.3}   & \add{8.2}  & \add{12.0}              \\
		\bottomrule 
	\end{tabular}%
	\label{tab:pt}%
\end{table*}%

\subsection{Ablation Study}
\textit{\textbf{ABS-REF efficacy:}} 
In Tab.\ref{tab:abs-ref}, we establish the contribution of the REF stage in our HPENet V2, which follows the separable ABS-REF paradigm with non-local MLPs. By removing the REF stage, HPENet V2 degenerates into a single-stage method, which we refer to as HPENet-dv in the table. We chose HPENet-dv as the baseline and expanded it by adding a REF stage behind each ABS stage, resulting in a 7.3\% mIoU performance gain. Further application of PE schemes to HPENet-dv yields a performance gain of 7.9\% mIoU, demonstrating that positional encoding is crucial for local aggregation. Finally, replacing the traditional PE with the proposed HPE achieves a performance gain of 1.4\% mIoU. These results confirm that while PE can introduce geometric information into local aggregation, it limits performance due to the restricted dimensionality of positional information. To verify the impact of parameters, we removed the grouping operation in the REF stage of HPENet-dv (32, [1, 1, 1, 1]), resulting in a model with only the ABS stage and the same number of parameters, termed HPENet-dv${}^{\star}$. However, HPENet-dv${}^{\star}$ only achieves 53.2\% mIoU, significantly underperforming HPENet-dv(32, [1, 1, 1, 1]). These results validate that the REF stage is a crucial component in our ABS-REF framework and that our HPE effectively supports this framework.

\vspace{0.7mm}
\textit{\textbf{More analyses on positional encoding:}} 
In Tab.~\ref{tab:hpe}, we evaluate the influence of different positional encodings at different stages of HPENet V2-XL on S3DIS Area-5. We compare the proposed High-dimensional Positional Encoding (HPE) with a conventional learnable positional encoding (PE). To study the effect of absolute coordinates, we feed absolute point positions into HPE, denoted HPE (abs). We also replace the standard element-wise addition used to fuse positional cues with element-wise multiplication, denoted HPE (mul), which treats the positional signal as a dynamic feature weight. \add{Note that relative positional encodings are translation invariant, whereas absolute encodings are not.}
Below, we summarize interesting observations regarding our results in Tab.~\ref{tab:hpe}. 
\textbf{Firstly}, applying HPE in both the ABS and REF stages consistently outperforms using it in either stage alone. This supports our claim that positional encoding is beneficial for both ABS and REF stages, which is a distinctive property of our framework. 
\textbf{Secondly}, HPE (abs) incurs 17.6\% mIoU absolute performance reductions, as compared to HPE. It suggests that absolute positional encoding is unsuitable for 3D semantic segmentation tasks since it can not capture local context information which is crucial for point cloud processing. 
\textbf{Thirdly}, HPE (mul) incurs 1.12\% mIoU reduction as compared to HPE. 
Therefore, we use element-wise addition to fuse positional encoding into features. 

\begin{figure}[t]
	\begin{center}
		\includegraphics[width=1.0\linewidth]{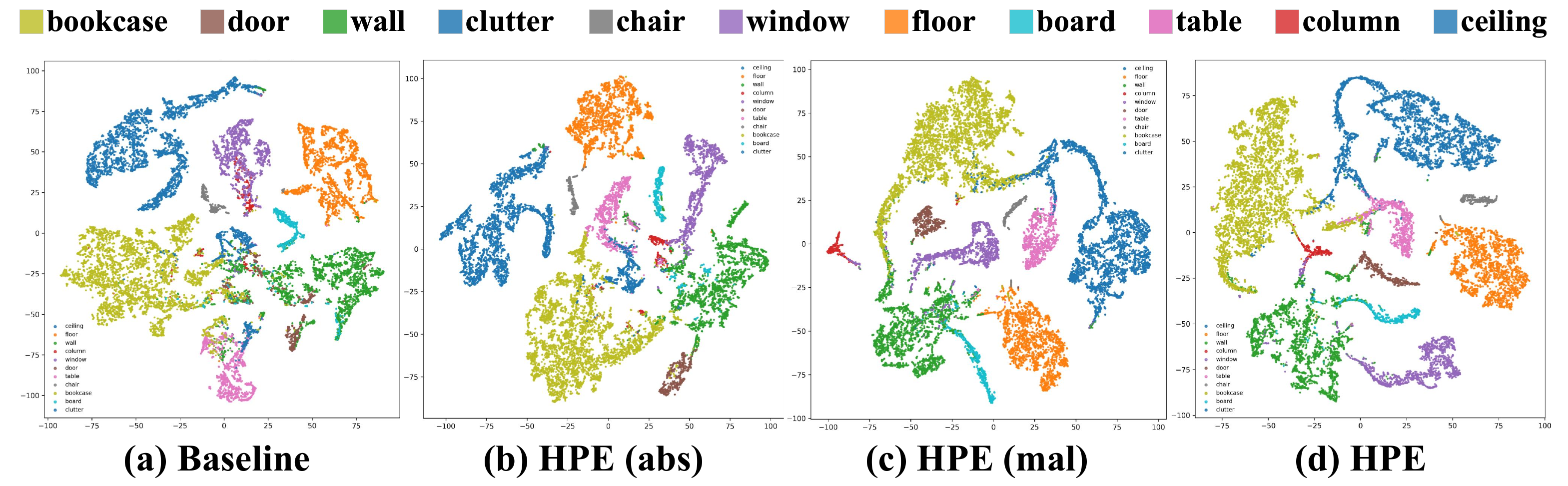}	
	\end{center}
	\vspace{-2mm}
	\caption{\add{t-SNE comparison of positional encoding variants for HPENet-dv. Left to right: (a) Baseline, (b) HPE (abs), (c) HPE (mal), (d) HPE.}}	
	\label{fig:positional}
	\vspace{-3mm}
\end{figure}

\add{For a deeper analysis of positional encoding variants, Fig.~\ref{fig:positional} shows t-distributed stochastic neighbor embedding (t-SNE) visualizations of the final-layer features from HPENet-dv. Relative to the baseline, HPE (abs) still yields fragmented clusters with noticeable class overlap, indicating weak discriminative power. HPE (mul) improves local cohesion but often forms elongated clusters with scattered islands. In contrast, the proposed HPE produces compact intra-class groupings, wider inter-class margins, and fewer outliers with cleaner boundaries. These patterns indicate that encoding relative coordinates reduces inter-class similarity and that additive fusion further strengthens within-class compactness, making the proposed HPE the most effective variant.}

In Tab.~\ref{tab:hpe-c}, we further analyze the effects of dimension variation of high-dimensional projected space with HPE on S3DIS Area-5. The results indicate that high-dimensional representation is crucial for positional encoding. When the feature dimension of high-dimensional representation increases from 3 to $C/4$, the network performance continues to improve, but when the feature dimension reaches over $C/4$, the performance tends to saturate. These results are also consistent with common sense that network performance tends to saturate with an increase in network width. Therefore, we set the dimension to $C/4$ in HPENet V2.

\vspace{0.7mm}
\textit{\textbf{Backward Fusion Module:}} 
In Tab.~\ref{tab:hpe}, we evaluate the influence of the proposed Backward Fusion Module (BFM). As shown in lines 4 and 5 in Tab.~\ref{tab:hpe}, HPENet V2 achieves a 0.5\% mAcc performance gain by further introducing BFM, indicating that embedding contextual information can improve the performance balance across categories in HPENet and enhance its generalization.

\begin{figure}[t]
	\begin{center}
		\includegraphics[width=0.9\linewidth]{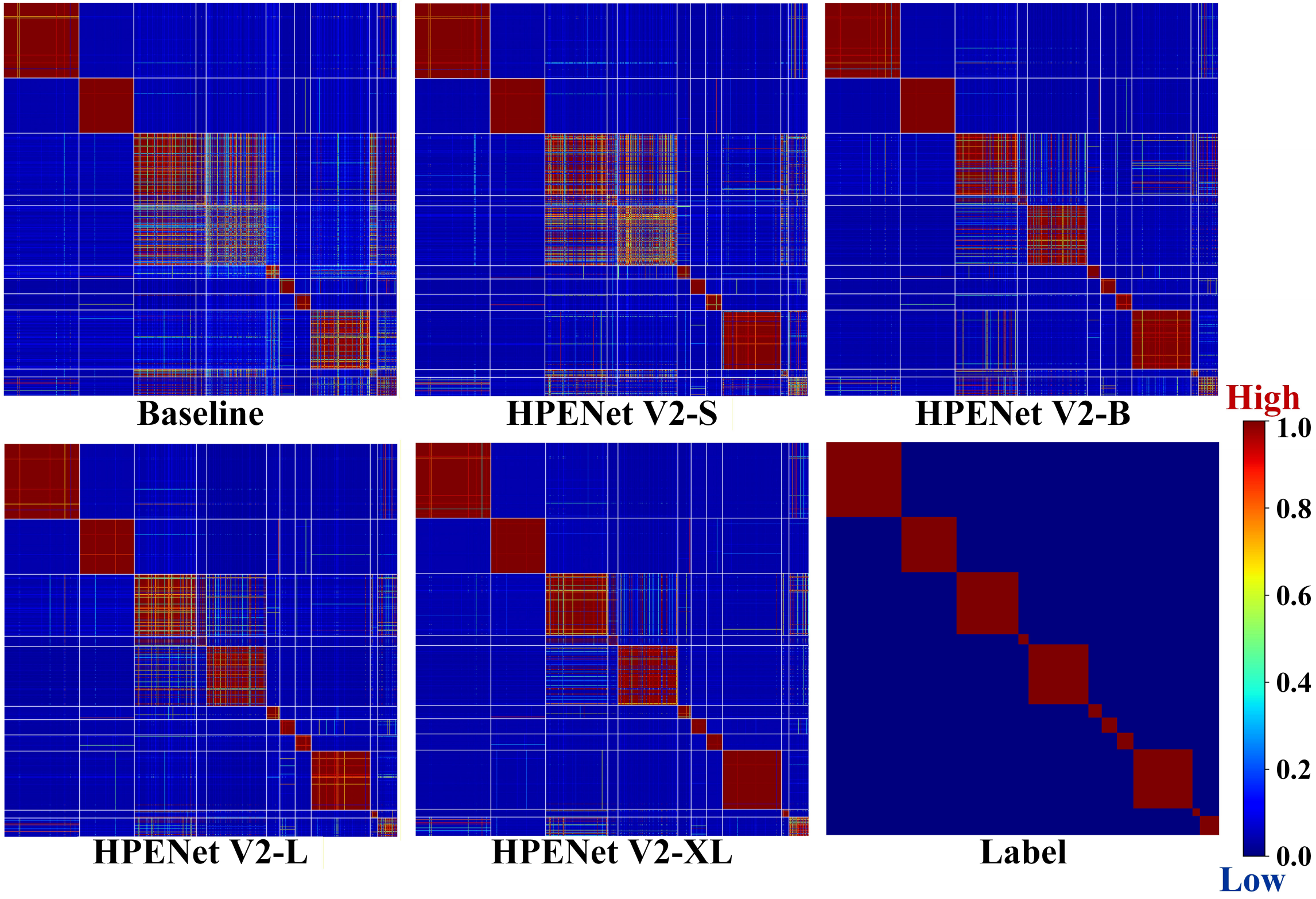}
	\end{center}
	\vspace{-3mm}
	\caption{\add{Visual comparison of feature-to-label similarity matrices for HPENet V2 variants. The baseline and label matrix are shown for reference.}}
	\label{fig:similarity}
	\vspace{-3mm}
\end{figure}

\vspace{0.7mm}
\textit{\textbf{\add{Different configurations of HPENet V2:}}}
\add{In Fig.~\ref{fig:similarity}, we visualize feature-to-label similarity matrices for HPENet V2 variants with varying depth and width. The baseline and a label-to-label matrix are included for reference. Each matrix is computed using cosine similarity between final-layer point features and one-hot class vectors. Rows are ordered by the predicted categories, and columns follow the class index order. 
The results indicate that introducing the proposed HPE and BFM (baseline vs.\ HPENet~V2-S) significantly enhances feature separability. Moreover, increasing the network depth and width further yields a sharper block-diagonal structure and lower off-diagonal responses in the similarity matrix compared with the baseline, reflecting stronger feature discrimination and improved segmentation performance.}

\subsection{When HPENet Meets Transformer-based Methods}
\label{sec:pt}
The above experimental results indicate that the proposed High-dimensional Positional Encoding (HPE), non-local MLPs, and Backward Fusion Module (BFM) substantially improve MLP-based models. \add{Recently, transformer-based methods have caused a paradigm shift in point cloud processing. As shown in Fig.~\ref{fig:ABS-REF}, the traditional transformer-based method, Point Transformer (PT)~\cite{3zhao2021point}, also follows the proposed ABS-REF framework. Stratified Transformer (ST)~\cite{6lai2022stratified} follows the same perspective. 
Therefore, we further integrate the proposed modules into Transformer-based methods and evaluate their effectiveness. 
Results on S3DIS Area-5 for PT and ST are reported in Tab.~\ref{tab:pt}, and detailed results are provided in the supplementary material~IV.}

\add{Tab.~\ref{tab:pt} shows all proposed modules achieve significant gains over PT and ST. First, replacing the traditional local MLPs with non-local MLPs in the ABS stage reduces FLOPs. While a reduction in FLOPs is expected, it is surprising that the performance improves by 1.5\% and 0.6\% mIoU for PT and ST, respectively. We attribute this to the Transformer layers in the REF stage capturing abundant contextual information from neighboring point sets, and the non-local MLP in the ABS stage providing complementary non-local updates across input point sets.
We next evaluate the proposed HPE. As shown in Tab.~\ref{tab:pt}, adding HPE to the modified PT yields a further 0.6\% mIoU improvement. This result also indicates that projecting the relative point coordinates into high-dimensional spaces is crucial. Finally, incorporating BFM to model contextual information yields a further 0.4\% mIoU gain.
Overall, incorporating our modules improves PT by 2.5\% mIoU while reducing compute by 0.4G FLOPs, and improves ST by 1.3\% mIoU with a 0.3G FLOP reduction. These results demonstrate the compatibility of our modules with Transformer-based methods.}

\subsection{\add{Robustness test}}
\begin{figure}[t]
	\begin{center}
		\includegraphics[width=1.0\linewidth]{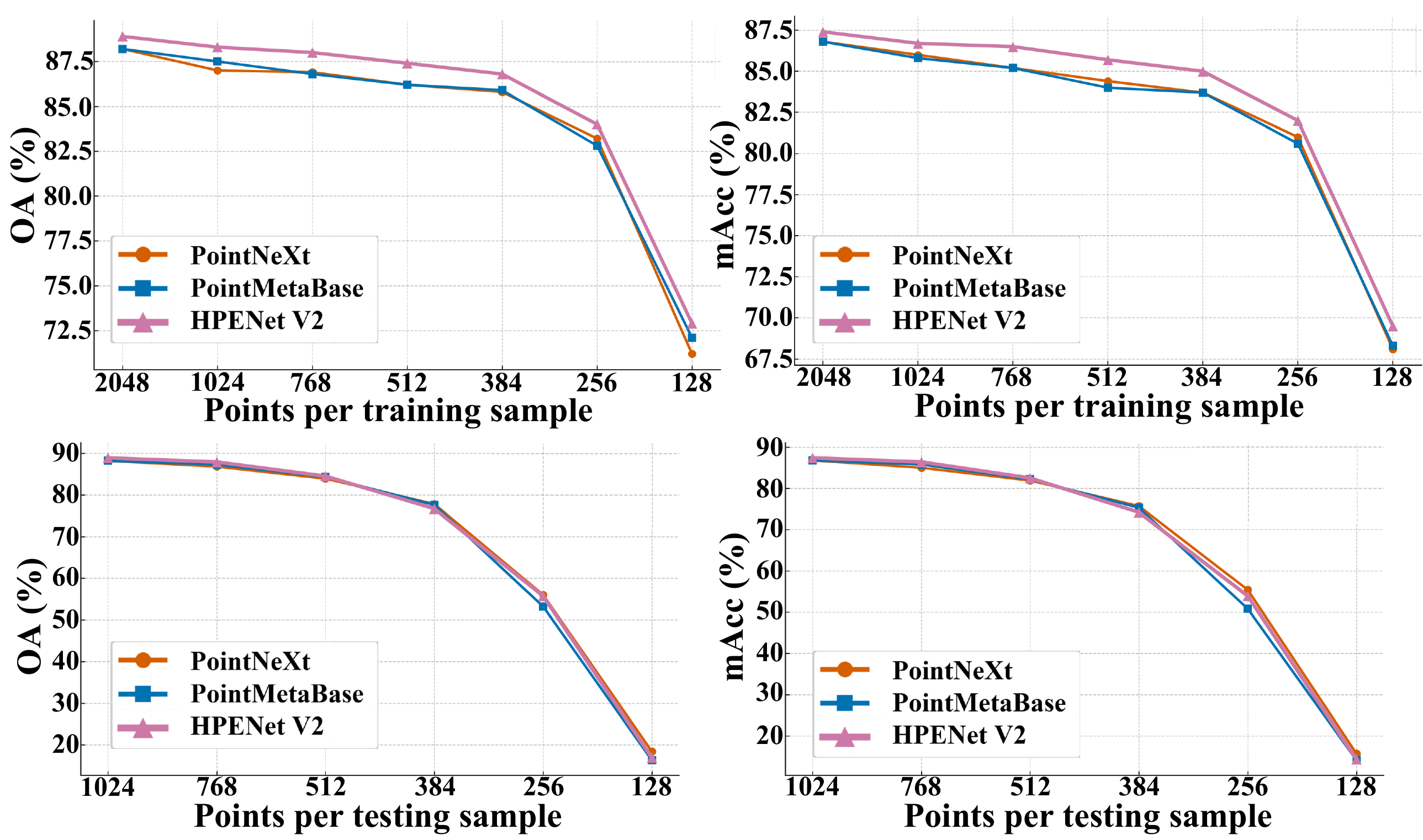}
	\end{center}
	\caption{\add{Robustness to point density during training and testing on ScanObjectNN across PointNeXt, PointMetaBase, and HPENet V2.}}
	\label{fig:density}
	\vspace{-3mm}
\end{figure}

\add{We further evaluate the robustness of HPENet V2 to point cloud density on ScanObjectNN~\cite{42uy2019revisiting} . PointNeXt~\cite{5qian2022pointnext} and PointMetaBase~\cite{98lin2023meta} serve as baselines, where PointNeXt uses local MLPs for local aggregation, whereas PointMetaBase employs only non-local MLPs. Except for the number of points, all configurations are identical. We use farthest point sampling to generate inputs with a range of point counts for both training and testing. As shown in the first row of Fig.~\ref{fig:density}, HPENet V2 maintains strong accuracy across densities and remains robust to sparse inputs. With only 256 points per sample during training, HPENet V2 achieves 84.0\% OA and 82.0\% mAcc. Under conditions with missing points, HPENet V2 also shows robust performance, which we attribute to the presence of local MLPs in the first ABS stage. These results indicate that local MLPs are resilient to variations in point cloud density and that using them in the first ABS stage improves the robustness of HPENet V2.}
      
\section{Conclusion}
Inspired by the distinct subsampling and convolution stages in image processing models, we provide a two-stage abstraction and refinement (ABS-REF) view for point cloud neural processing. This view offers an intuitive way to highlight the key strengths of existing methods. 
Additionally, we propose a High-dimensional Positional Encoding (HPE) scheme that effectively captures local geometric information, which is compatible with the ABS-REF paradigm. Furthermore, we design a simple and effective Backward Fusion Module (BFM) to embed context information into the processing pipeline.
Within our ABS-REF framework, we rethink the local aggregation in MLP-based methods. We propose removing expensive set operations in local aggregation and introducing non-local MLPs to process input point sets directly. To leverage the advantages and potential of various local aggregation, we propose to use different local aggregation in the ABS and REF stages.
With the proposed ABS-REF view, HPE, and BFM, we have devised a suite of HPENets that leverage HPE for MLP-based modeling in object classification, object part segmentation, semantic segmentation, and object detection, largely improving SOTA performance across the board. 
Extensive experiments demonstrate that HPENets not only achieve impressive performance but are also highly efficient. 
\add{Moreover, our proposed modules can enhance MLP-based architectures and are also compatible with Transformer-based methods, such as Point Transformer and Stratified Transformer.}

\ifCLASSOPTIONcompsoc
\section*{Acknowledgments}
\else
\section*{Acknowledgment}
\fi
This work was supported by National Natural Science Foundation of China under Grant (62373140, U21A20487, U2013203), National Key R\&D Program (2023YFB4704500), Leading Talents in Science and Technology Innovation of Hunan Province (2023RC1040), the Project of Science Fund of Hunan Province (2022JJ30024); the Project of Talent Innovation and Sharing Alliance of Quanzhou City (2021C062L).
%

%
\bibliography{ref}

\clearpage
\setcounter{section}{0}            
\renewcommand{\thesection}{\Roman{section}}

\setcounter{subsection}{0}
\renewcommand{\thesubsection}{\thesection.\Alph{subsection}}
\section*{APPENDIX}
This supplementary material provides additional content to complement the main manuscript and is divided into the following sections.

\begin{itemize}
	\item The architecture of HPENet V2 for 3D object classification.
	\item \add{Theoretical analysis of Abstraction and Refinement (ABS-REF) View.}
	\item \add{Theoretical analysis of High-dimensional Positional Encoding.}
	\item Details of experimental results on S3DIS for indoor scene semantic segmentation.
	\item \add{Details of experimental results on SemanticKITTI for outdoor scene semantic segmentation.}
	\item More instantiations of the proposed ABS-REF framework.
	\item Limitation and future work.
	\item More qualitative comparisons on S3DIS and ScanNet V2.
\end{itemize}

Note that the supplementary material will be made public through arXiv after acceptance.

\section{3D Object Classification Architecture}
\begin{figure}[htbp]
	\begin{center}
		\includegraphics[width=1\linewidth]{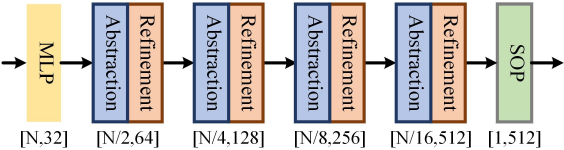}
	\end{center}
	\caption{\textbf{3D object classification architecture of HPENet V2.} The classification architecture has the same encoder structure as the semantic segmentation model (Fig.~2 in the manuscript).
	}
	\label{fig:cls}
\end{figure}

Fig.~\ref{fig:cls} illustrates the classification architecture of HPENet V2. As shown in Fig.~\ref{fig:cls}, the classification architecture has the same encoder structure as the semantic segmentation model (Fig.~2 in the manuscript). However, since the input points are typically small in classification and part segmentation tasks (e.g., 1024 or 2048 points), the points are downsampled only twice, compared to four times in scene semantic segmentation tasks. A symmetric operation is then applied to aggregate the output features of the encoder into a global shape representation for classification tasks. 

\begin{figure*}[t]
	\begin{center}
		\includegraphics[width=0.9\textwidth]{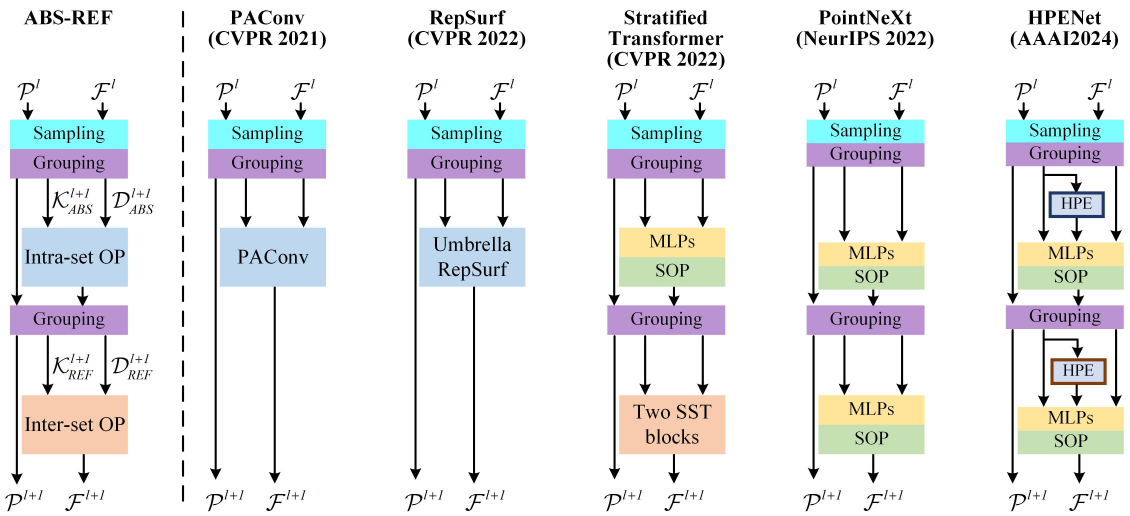}
	\end{center}
	\caption{\textbf{More recent representative instantiations of the ABS-REF framework.} Abbreviations include SOP: Symmetric OPeration, OP: aggregation OPeration, Two SST blocks: Two Successive Stratified Transformer Blocks.
	}
	\label{fig:hpenet_ins}
\end{figure*}

\begin{table*}[htbp]
	\centering
	\caption{Detailed semantic segmentation results (\%) on S3DIS~\cite{43armeni20163d} with 6-fold cross-validation.}
	
	\setlength\tabcolsep{2pt}
	\begin{tabular}{@{}lccccccccccccccccc@{}}
		\toprule
		Methods & Area  & OA    & mACC  & mIoU  & Ceiling & Floor & Wall  & Beam  & Column & Window & Door  & Chair & Table & Bookcase & Sofa  & Board & Clutter \\
		\midrule
		\multirow{7}[0]{*}{HPENet V2-S} & 1     & 90.8  & 85.9  & 77.4  & 96.3  & 96.0  & 84.5  & 64.9  & 60.9  & 81.4  & 88.0  & 73.2  & 83.0  & 74.5  & 61.7  & 71.8  & 70.3  \\
		& 2     & 81.9  & 66.2  & 51.9  & 90.4  & 86.9  & 80.1  & 9.0   & 44.5  & 17.5  & 63.4  & 49.6  & 43.5  & 69.3  & 49.7  & 30.0  & 41.1  \\
		& 3     & 91.4  & 85.2  & 76.5  & 96.3  & 98.2  & 84.4  & 66.8  & 60.5  & 48.8  & 87.7  & 72.2  & 81.5  & 74.0  & 68.5  & 82.1  & 73.4  \\
		& 4     & 86.9  & 71.7  & 60.9  & 92.0  & 97.6  & 82.5  & 9.0   & 45.7  & 36.5  & 68.2  & 66.3  & 76.1  & 51.6  & 56.2  & 52.7  & 57.9  \\
		& 5     & 89.4  & 72.4  & 65.6  & 93.6  & 98.5  & 83.4  & 0.0   & 32.3  & 52.7  & 68.0  & 81.7  & 88.8  & 57.9  & 72.4  & 66.2  & 57.4  \\
		& 6     & 93.5  & 90.7  & 83.6  & 96.7  & 97.9  & 88.2  & 77.7  & 77.2  & 86.3  & 91.2  & 79.7  & 88.2  & 72.4  & 76.2  & 79.3  & 75.8  \\
		\cmidrule{2-18}
		& \textbf{6-fold}   & 88.7  & 79.6  & 70.5  & 93.8  & 95.5  & 83.6  & 59.5  & 52.7  & 58.4  & 75.3  & 74.1  & 68.7  & 65.6  & 66.1  & 64.1  & 59.1  \\
		\midrule
		\multirow{7}[0]{*}{HPENet V2-B} & 1     & 91.4  & 87.0  & 79.5  & 96.5  & 96.2  & 83.9  & 59.9  & 68.1  & 87.1  & 90.3  & 74.3  & 87.1  & 79.9  & 64.2  & 71.5  & 74.7  \\
		& 2     & 89.0  & 77.7  & 66.5  & 91.3  & 95.8  & 84.7  & 40.7  & 69.5  & 74.6  & 74.9  & 57.7  & 71.6  & 68.1  & 56.9  & 20.1  & 58.5  \\
		& 3     & 93.4  & 92.3  & 83.7  & 95.9  & 98.5  & 87.9  & 75.0  & 54.6  & 88.8  & 92.2  & 76.6  & 86.7  & 89.2  & 74.5  & 90.2  & 77.9  \\
		& 4     & 89.8  & 81.8  & 69.4  & 95.7  & 98.0  & 84.6  & 44.4  & 66.5  & 38.6  & 71.2  & 66.1  & 83.6  & 62.5  & 66.1  & 57.5  & 66.8  \\
		& 5     & 90.4  & 76.5  & 70.1  & 94.7  & 98.6  & 83.6  & 0.0   & 35.9  & 56.8  & 74.8  & 83.1  & 91.5  & 76.5  & 77.2  & 76.9  & 61.7  \\
		& 6     & 94.4  & 92.3  & 86.2  & 97.0  & 98.0  & 90.2  & 84.5  & 83.1  & 89.0  & 91.7  & 80.8  & 88.8  & 82.9  & 77.6  & 80.7  & 76.9  \\
		\cmidrule{2-18}
		& \textbf{6-fold}   & 91.0  & 84.3  & 76.0  & 94.9  & 97.5  & 85.2  & 61.0  & 60.7  & 67.3  & 80.3  & 75.9  & 81.7  & 75.5  & 71.8  & 68.3  & 67.5  \\
		\midrule
		\multirow{7}[0]{*}{HPENet V2-L} & 1     & 92.0  & 87.9  & 80.5  & 96.9  & 96.3  & 85.6  & 71.0  & 69.7  & 85.9  & 86.5  & 74.1  & 85.6  & 81.5  & 63.4  & 74.6  & 75.3  \\
		& 2     & 90.0  & 77.5  & 67.4  & 90.3  & 97.5  & 85.7  & 41.2  & 68.9  & 81.8  & 83.4  & 65.5  & 77.7  & 52.1  & 58.9  & 14.7  & 59.0  \\
		& 3     & 93.6  & 91.5  & 84.3  & 96.0  & 98.5  & 88.0  & 76.4  & 55.2  & 89.7  & 93.2  & 78.9  & 87.2  & 90.7  & 75.2  & 88.9  & 77.6  \\
		& 4     & 90.4  & 83.5  & 72.2  & 96.1  & 98.2  & 85.1  & 55.9  & 75.4  & 41.5  & 76.3  & 70.3  & 82.3  & 66.6  & 65.8  & 58.1  & 67.2  \\
		& 5     & 91.2  & 77.3  & 71.1  & 95.2  & 98.6  & 84.9  & 0.0   & 42.5  & 56.9  & 74.4  & 81.8  & 90.5  & 74.8  & 80.2  & 80.0  & 64.8  \\
		& 6     & 94.3  & 92.2  & 86.7  & 97.1  & 98.0  & 90.0  & 83.6  & 81.2  & 89.4  & 92.3  & 79.9  & 87.2  & 92.5  & 76.4  & 83.2  & 76.7  \\
		\cmidrule{2-18}
		& \textbf{6-fold}   & 91.6  & 85.3  & 77.5  & 95.0  & 97.9  & 86.1  & 68.1  & 64.4  & 67.9  & 82.7  & 76.9  & 83.3  & 74.3  & 73.0  & 69.3  & 68.5  \\
		\midrule
		\multirow{7}[0]{*}{HPENet V2-XL} & 1     & 92.3  & 88.7  & 81.7  & 96.6  & 96.1  & 86.2  & 74.0  & 70.4  & 84.5  & 91.4  & 74.4  & 87.8  & 82.7  & 65.4  & 77.7  & 75.0  \\
		& 2     & 89.6  & 77.9  & 69.0  & 92.5  & 96.8  & 84.5  & 46.3  & 74.8  & 82.7  & 76.8  & 57.4  & 74.1  & 78.9  & 54.6  & 18.2  & 59.8  \\
		& 3     & 94.2  & 92.6  & 86.0  & 96.3  & 98.6  & 89.3  & 80.1  & 65.8  & 91.4  & 93.4  & 78.2  & 87.8  & 91.5  & 76.3  & 90.7  & 78.0  \\
		& 4     & 91.2  & 83.7  & 72.8  & 95.8  & 98.4  & 87.4  & 43.6  & 77.5  & 47.3  & 75.9  & 73.1  & 83.9  & 72.9  & 68.0  & 53.3  & 68.9  \\
		& 5     & 91.6  & 78.8  & 72.6  & 94.7  & 98.6  & 86.8  & 0.0   & 44.2  & 61.3  & 73.9  & 83.2  & 92.1  & 82.4  & 79.6  & 83.8  & 62.9  \\
		& 6     & 94.5  & 92.7  & 87.5  & 97.1  & 97.9  & 91.1  & 86.8  & 87.9  & 87.5  & 92.3  & 79.3  & 89.1  & 92.0  & 77.3  & 82.5  & 76.0  \\
		\cmidrule{2-18}
		& \textbf{6-fold}   & 91.9  & 86.3  & 78.9  & 95.2  & 97.7  & 87.2  & 71.0  & 68.3  & 69.9  & 81.9  & 76.9  & 82.9  & 82.7  & 73.1  & 71.0  & 68.4  \\
		\midrule
		Point Transformer~\cite{3zhao2021point} & 5 & 90.4  & 74.8  & 68.2  & 93.9  & 98.5  & 85.8  & 0.1   & 33.4  & 61.4  & 74.6  & 82.0  & 90.7  & 65.7  & 74.4  & 67.1  & 59.1 \\
		+Non-local MLPs & 5 &90.7  & 76.3  & 69.7  & 94.7  & 97.6  & 86.4  & 0.0   & 38.7  & 62.0  & 75.0  & 80.2  & 90.3  & 73.2  & 72.7  & 75.2  & 60.1  \\
		+HPE & 5 & 90.5  & 76.4  & 70.3  & 93.5  & 98.3  & 86.1  & 0.0   & 36.2  & 63.4  & 78.3  & 81.8  & 91.5  & 76.9  & 73.0  & 77.0  & 57.8  \\
		+BFM & 5 &90.9  & 76.6  & 70.7  & 94.3  & 98.5  & 86.7  & 0.0   & 36.7  & 62.1  & 77.2  & 82.4  & 89.2  & 76.0  & 74.6  & 81.7  & 59.3  \\
		\midrule
		\add{Stratified Transformer~\cite{6lai2022stratified}} &
		\add{5} & \add{91.7} & \add{77.4} & \add{71.4} & \add{95.3} & \add{98.8} & \add{86.8} & \add{0.0}  & \add{45.8} & \add{63.0} & \add{73.8} & \add{84.6} & \add{91.2} & \add{68.3} & \add{78.3}  & \add{78.5} & \add{63.3}\\
		\add{+Non-local MLPs} &
		\add{5} & \add{90.9} & \add{78.9} & \add{72.0} &
		\add{95.6} & \add{98.6} & \add{84.6} & \add{0.0} &
		\add{52.5} & \add{62.9} & \add{72.6} & \add{84.2} &
		\add{91.1} & \add{72.7} & \add{77.1} & \add{79.5} & \add{64.3} \\
		\add{+HPE} &
		\add{5} & \add{92.2} & \add{78.7} & \add{72.5} &
		\add{95.3} & \add{98.7} & \add{87.7} & \add{0.0} &
		\add{48.6} & \add{62.0} & \add{77.7} & \add{85.6} &
		\add{91.1} & \add{68.7} & \add{80.5} & \add{79.9} & \add{66.2} \\
		\add{+BFM} &
		\add{5} & \add{91.9} & \add{78.7} & \add{72.7} &
		\add{95.0} & \add{98.8} & \add{87.6} & \add{0.0} &
		\add{53.4} & \add{63.3} & \add{72.8} & \add{84.7} &
		\add{91.3} & \add{74.2} & \add{78.6} & \add{81.6} & \add{64.3} \\
		\bottomrule  
	\end{tabular}
	\label{tab:s3dis 6-fold}%
\end{table*}%
\begin{table*}[htbp]
	\centering
	\caption{Detailed semantic segmentation results (\%) on SemanticKITTI.}
	
	\setlength\tabcolsep{4pt}
	\begin{tabular}{@{}lcccccccccccccccccccc@{}}
	\toprule
	\rotatebox{-90}{Method} &
	\rotatebox{-90}{mIoU} &
	\rotatebox{-90}{car} &
	\rotatebox{-90}{bicycle} &
	\rotatebox{-90}{motorcycle} &
	\rotatebox{-90}{truck} &
	\rotatebox{-90}{other-vehicle} &
	\rotatebox{-90}{person} &
	\rotatebox{-90}{bicyclist} &
	\rotatebox{-90}{motorcyclist} &
	\rotatebox{-90}{road} &
	\rotatebox{-90}{parking} &
	\rotatebox{-90}{sidewalk} &
	\rotatebox{-90}{other-ground} &
	\rotatebox{-90}{building} &
	\rotatebox{-90}{fence} &
	\rotatebox{-90}{vegetation} &
	\rotatebox{-90}{trunk} &
	\rotatebox{-90}{terrain} &
	\rotatebox{-90}{pole} &
	\rotatebox{-90}{traffic-sign} \\
	\midrule
	\add{HPENet V2-S}  &
	\add{66.9}  & \add{95.5} & \add{49.9} & \add{79.3} &
	\add{85.7} & \add{52.2} & \add{75.3} & \add{90.8} & \add{3.6}  &
	\add{93.9} & \add{49.9} & \add{82.8} & \add{0.7}  & \add{90.0} &
	\add{62.8} & \add{90.0} & \add{72.7} & \add{79.4} & \add{65.5} & \add{50.4} \\
	\add{HPENet V2-B}  &
	\add{67.0} & \add{95.7} & \add{43.8} & \add{77.3} &
	\add{91.7} & \add{55.0} & \add{75.6} & \add{91.5} & \add{2.1}  &
	\add{94.0} & \add{49.0} & \add{82.7} & \add{6.8}  & \add{90.6} &
	\add{66.0} & \add{88.5} & \add{70.4} & \add{75.4} & \add{65.4} & \add{51.6} \\
	\add{HPENet V2-L}  &
	\add{67.4}  & \add{96.2} & \add{52.7} & \add{77.1} &
	\add{88.1} & \add{64.2} & \add{73.6} & \add{88.3} & \add{0.1}  &
	\add{94.0} & \add{54.9} & \add{82.0} & \add{0.4}  & \add{91.1} &
	\add{67.6} & \add{88.6} & \add{70.3} & \add{75.8} & \add{64.8} & \add{51.4} \\
	\add{HPENet V2-XL} &
	\add{69.3}  & \add{96.5}  & \add{55.3} & \add{78.8} &
	\add{92.8} & \add{66.2} & \add{78.4} & \add{86.1} & \add{25.2} &
	\add{94.6} & \add{48.6} & \add{83.9} & \add{0.3}  & \add{91.7} &
	\add{69.2} & \add{87.7} & \add{69.8} & \add{74.0} & \add{65.4} & \add{51.7} \\
	\bottomrule 
	\end{tabular}
	\label{tab:semantickitti}%
\end{table*}%

\section{Theoretical analysis of Abstraction and Refinement View}
\add{\subsection{ABS-REF view is Modular}}
Conventional point-cloud backbones often treat feature abstraction and feature refinement as loosely coupled steps within a single encoder. Although PointMetaBase proposes a general framework (PointMeta), it primarily emphasizes local aggregation and does not explicitly model the sampling operations that modify spatial resolution. Under our ABS–REF view, the encoder consists of an \emph{ABS} stage followed by an \emph{REF} stage. The ABS stage abstracts features from the input point cloud and produces a downsampled set of points via sampling and local aggregation operations. The REF stage refines these abstracted features using stacked resolution-preserving blocks to improve representation quality and scalability without changing the number of points.

\add{\subsection{ABS-REF view is clear}}
\add{Beyond disentangling the encoder, the ABS–REF view provides an intuitive lens to characterize the key strengths of existing backbones. As illustrated in Fig.~3 in the manuscript, early methods such as PointNet++~\cite{8qi2017pointnet++} and PointConv~\cite{20wu2019pointconv} consist only of the ABS stage. They employ different intra-set operators for local aggregation (MLPs in PointNet++ and density-aware discrete convolution in PointConv), yet they are single-stage models in our perspective and essentially lack a REF stage. More recent techniques report stronger performance by introducing an explicit REF stage. Within the ABS–REF view, Point Transformer~\cite{3zhao2021point} includes positional encoding together with Point Transformer layers as its refinement component, while PointMixer~\cite{4choe2022pointmixer} refines through the Mixer block. This interpretation clarifies that gains arise not only from better local aggregation in ABS but also from resolution-preserving refinement in REF that expands contextual reasoning and deepens the network.}

\add{\subsection{ABS-REF view is compatible}}
\add{The ABS–REF view further unifies diverse MLP- and Transformer-based point-cloud models under a single compatible paradigm. Modules designed for MLP backbones can be transferred to Transformer backbones when they are positioned according to their roles in ABS or REF. See Sec.~5.6 of the manuscript for empirical validation.}

\add{\section{Theoretical analysis of High-dimensional Positional Encoding}}

\add{\subsection{HPE is translation invariant.}}
\add{High-dimensional positional encoding (HPE) is constructed on \emph{relative} point positions, \(\Delta p_{mj}=p_j-p_m\).
Consequently, local operators that depend on HPE are insensitive to global translations of the point cloud, since a shift \(p_m \mapsto p_m+t\) leaves all \(\Delta p_{mj}\) unchanged.
This follows the symmetry principles in geometric deep learning on Euclidean domains~\cite{150bronstein2017geometric}, where point-cloud networks should be insensitive to global translations while still exploiting local geometric structure.}
	
\add{\subsection{The use of HPE is permutation invariance}}
\add{The use of HPE in HPENet~V2 can be written as
	\[
	f_m \;=\; \mathrm{MaxPool}_{j \in \mathcal{N}(m)} \big( f_j \;+\; \mathrm{HPE}(\Delta p_{mj}) \big),
	\]
	where \(f_j \in \mathbb{R}^C\) denotes the feature of a neighbor point \(j\),
	\(\mathrm{HPE}(\Delta p_{mj}) \in \mathbb{R}^C\) is the high-dimensional positional encoding computed from the relative coordinate \(\Delta p_{mj}=p_j-p_m\),
	and \(\mathrm{MaxPool}\) denotes max pooling over the neighbor dimension (aggregating an \(N \times K \times C\) tensor into \(N \times C\)).
	From a geometric deep learning perspective, this is a local message-passing operator in which the message from \(j\) to \(m\) is modulated by a learnable function of their relative position \(\Delta p_{mj}\).
	Using relative coordinates preserves translation structure, and the subsequent max pooling provides a \emph{symmetric}, permutation-invariant aggregation over the unordered neighbor set \(\mathcal{N}(m)\), consistent with standard set- and point-cloud frameworks~\cite{149zaheer2017deep}.}
	
\add{\subsection{Why higher dimensionality helps}
	Mapping input coordinates to higher-frequency (higher-dimensional) vectors and then passing them to an MLP can substantially improve the performance of coordinate-based MLPs.
	As shown in~\cite{148tancik2020fourier}, a Fourier-feature mapping alleviates the spectral bias of coordinate MLPs toward low frequencies, enabling the learning of higher-frequency signals and thus improving accuracy.
	Our \({HPE}_{SIN}\) follows this principle by first using sinusoidal functions to project 3D relative coordinates into a high-dimensional space, while \({HPE}_{MLP}\) learns such a high-dimensional geometric expansion via a shared MLP.}

\section{Detailed Results on S3DIS}
As mentioned in the manuscript, we perform the standard 6-fold cross-validation on S3DIS~\cite{43armeni20163d}. For each area, we use the remaining five areas as the training set. Specifically, Tab.~\ref{tab:s3dis 6-fold} shows our results for each area in S3DIS, including overall accuracy (OA), mean accuracy (mAcc), and mean Intersection-over-Union (mIoU) for the 13 semantic classes. \add{We not only provide detailed results for our HPENet V2, but also present the details of the ablation study on Point Transformer~\cite{3zhao2021point} and 	Stratified Transformer~\cite{6lai2022stratified}, as related to Tab.~13 in the manuscript.}

\add{\section{Detailed Results on SemanticKITTI}}
\add{In this section, we present detailed results on SemanticKITTI~\cite{137behley2019semantickitti}. Tab.~\ref{tab:semantickitti} reports per-class IoU together with OA, mAcc, and mIoU for the 19 semantic classes.}

\section{Instantiation of ABS-REF Framework}
In this section, we present more recent representative instantiations of the ABS-REF framework in Fig.~\ref{fig:hpenet_ins}, such as PAConv~\cite{21xu2021paconv}, RepSurf~\cite{54ran2022surface}, Stratified Transformer~\cite{6lai2022stratified}, and PointNeXt~\cite{5qian2022pointnext}.
In our ABS-REF view, PAConv and RepSurf are single-stage methods that only include the ABS stage, whereas the others are two-stage methods. The key distinction between the single-stage methods lies in the intra-set operations within the ABS stage. 
For instance, PAConv dynamically assembles basic weight matrices stored in a weight bank to create a flexible convolution kernel, while RepSurf introduces a novel explicit representation of local structure.
As for the two-stage methods, such as Stratified Transformer and PointNeXt, they focus on the inter-set OP sub-stage in the REF stage. For example, Stratified Transformer employs two successive Stratified Transformer blocks to capture long-range dependencies in point clouds. PointNeXt introduces an inverted residual bottleneck module as the inter-set OP to further enhance the scalability of PointNet++~\cite{8qi2017pointnet++}.

\section{Limitation and Future Work}
\noindent\textit{\textbf{Limitation analysis.}} HPENet V2 demonstrates both efficient and impressive performance on real-world datasets, particularly S3DIS~\cite{43armeni20163d} and ScanObjectNN~\cite{42uy2019revisiting}. However, it may be suboptimal for synthetic object classification on ModelNet40~\cite{41wu20153d} and ShapeNetPart~\cite{53yi2016scalable}, as the point distributions are similar, making it challenging to capture geometric information from relative point coordinates using high-dimensional positional encoding (HPE).

HPENets are MLP-based methods that achieve state-of-the-art performance on S3DIS~\cite{43armeni20163d} while yielding comparable results on ScanNet~\cite{44dai2017scannet}. The advantages of HPENet stem from the powerful geometric representation provided by HPE, while its limitations arise from the lack of long-range dependency modeling due to the simple feature aggregation operations used in HPENet, such as max pooling.

\vspace{0.7mm}
\noindent\textit{\textbf{Future work.}} We plan to extend HPENet~V2 to broader application scenarios, including industrial settings, and to additional 3D point cloud processing tasks such as 3D instance segmentation.

\section{Qualitative results}
\begin{figure*}[t]
	\begin{center}
		\includegraphics[width=0.95\textwidth]{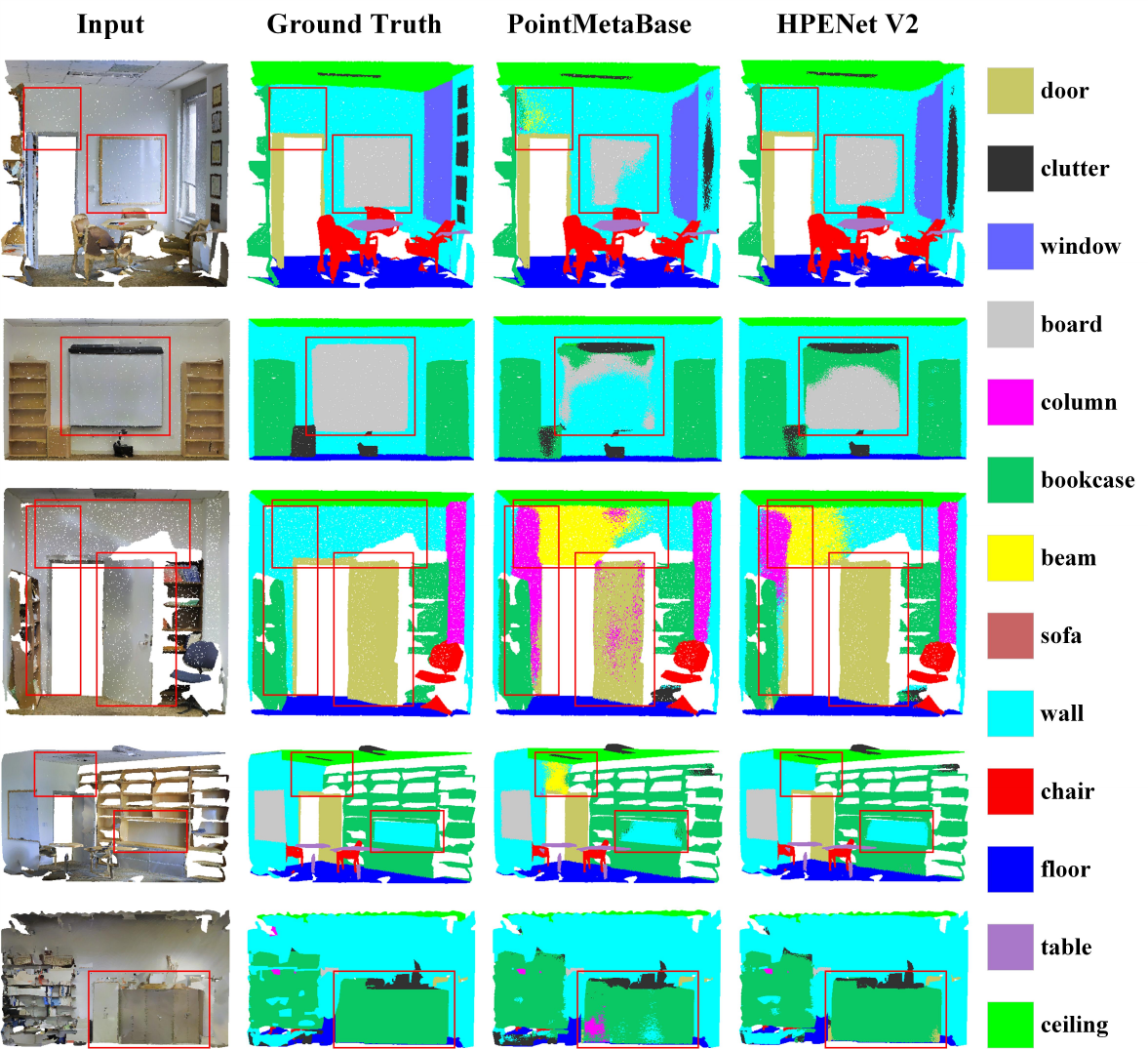}
	\end{center}
	\caption{Representative qualitative results of HPENet V2 and the balanced MLP-based method PointMetaBase~\cite{98lin2023meta} on S3DIS Area-5. The main differences are highlighted in red frames.
	}
	\label{fig:s3disv1}
\end{figure*}

\begin{figure*}[t]
	\begin{center}
		\includegraphics[width=0.95\textwidth]{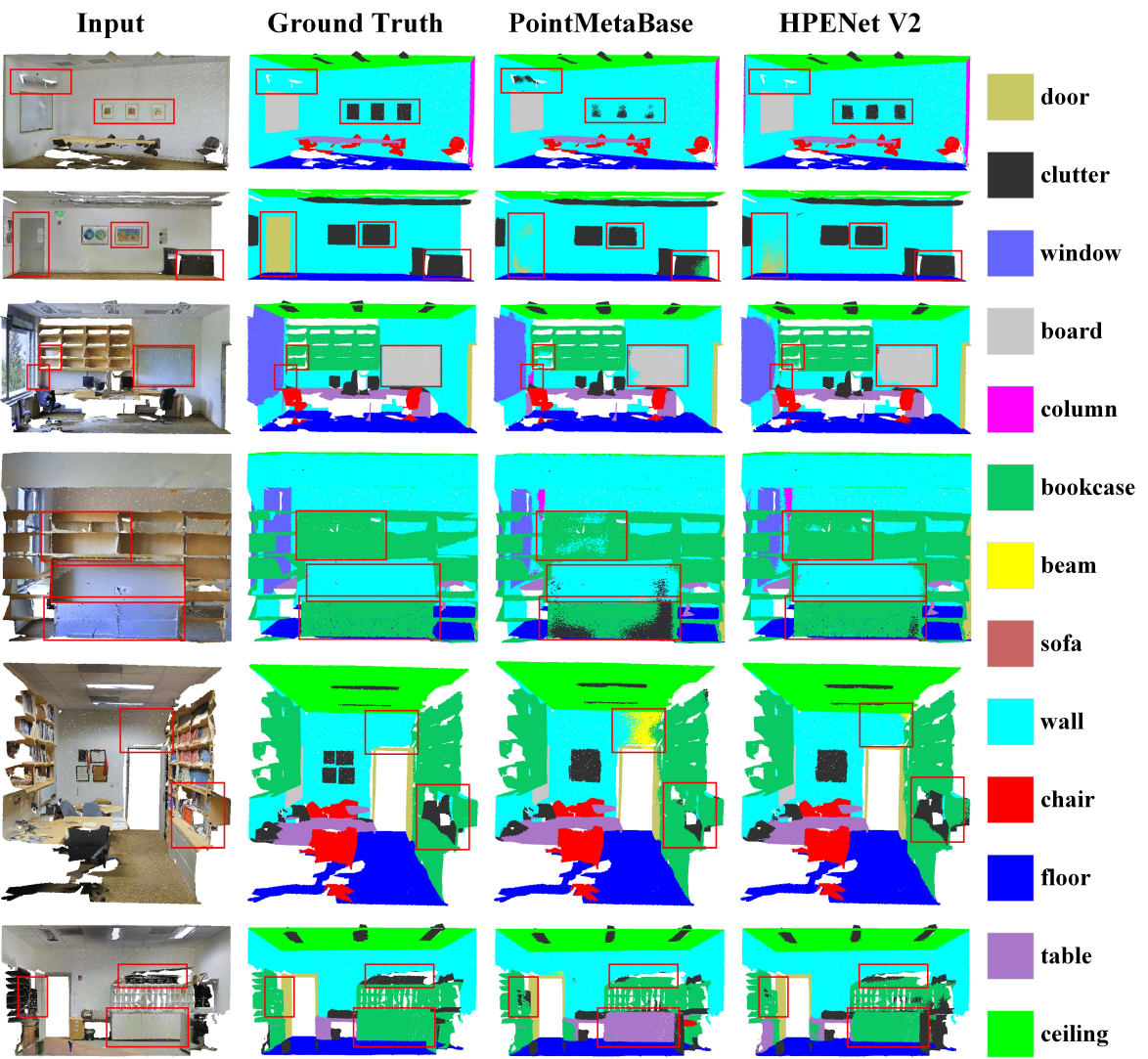}
	\end{center}
	\caption{Representative qualitative results of HPENet V2 and the balanced MLP-based method PointMetaBase~\cite{98lin2023meta} on S3DIS Area-5. The main differences are highlighted in red frames.
	}
	\label{fig:s3disv2}
\end{figure*}

\noindent\textit{\textbf{More qualitative results on S3DIS.}} 
Fig.~\ref{fig:s3disv1} and Fig.~\ref{fig:s3disv2} show additional qualitative results produced by HPENet V2 on S3DIS. As seen in Fig.~\ref{fig:s3disv1}, HPENet V2 provides predictions closer to the ground truth compared to the balanced MLP-based method PointMetaBase~\cite{98lin2023meta}. Specifically, HPENet V2 successfully segments the wall ($1^{st}$, $3^{rd}$, and $4^{th}$ rows), board ($1^{st}$ and $2^{nd}$ rows), door ($3^{rd}$ row) and bookcase ($5^{th}$ row), whereas PointMetaBase fails to segment these objects accurately in some cases.

Fig.~\ref{fig:s3disv2} further highlights the advantages of HPENet V2 on S3DIS compared to PointMetaBase. For example, HPENet V2 is able to segment wall ($1^{st}$, $3^{rd}$, $4^{th}$, and $5^{th}$ rows), clutter ($1^{st}$, $2^{nd}$, $5^{th}$ and $6^{th}$ rows), bookcase ($4^{th}$ and $6^{th}$ rows), and door ($2^{nd}$ row), while PointMetaBase fails to segment appropriately to some extent.

\begin{figure*}[t]
	\begin{center}
		\includegraphics[width=0.9\textwidth]{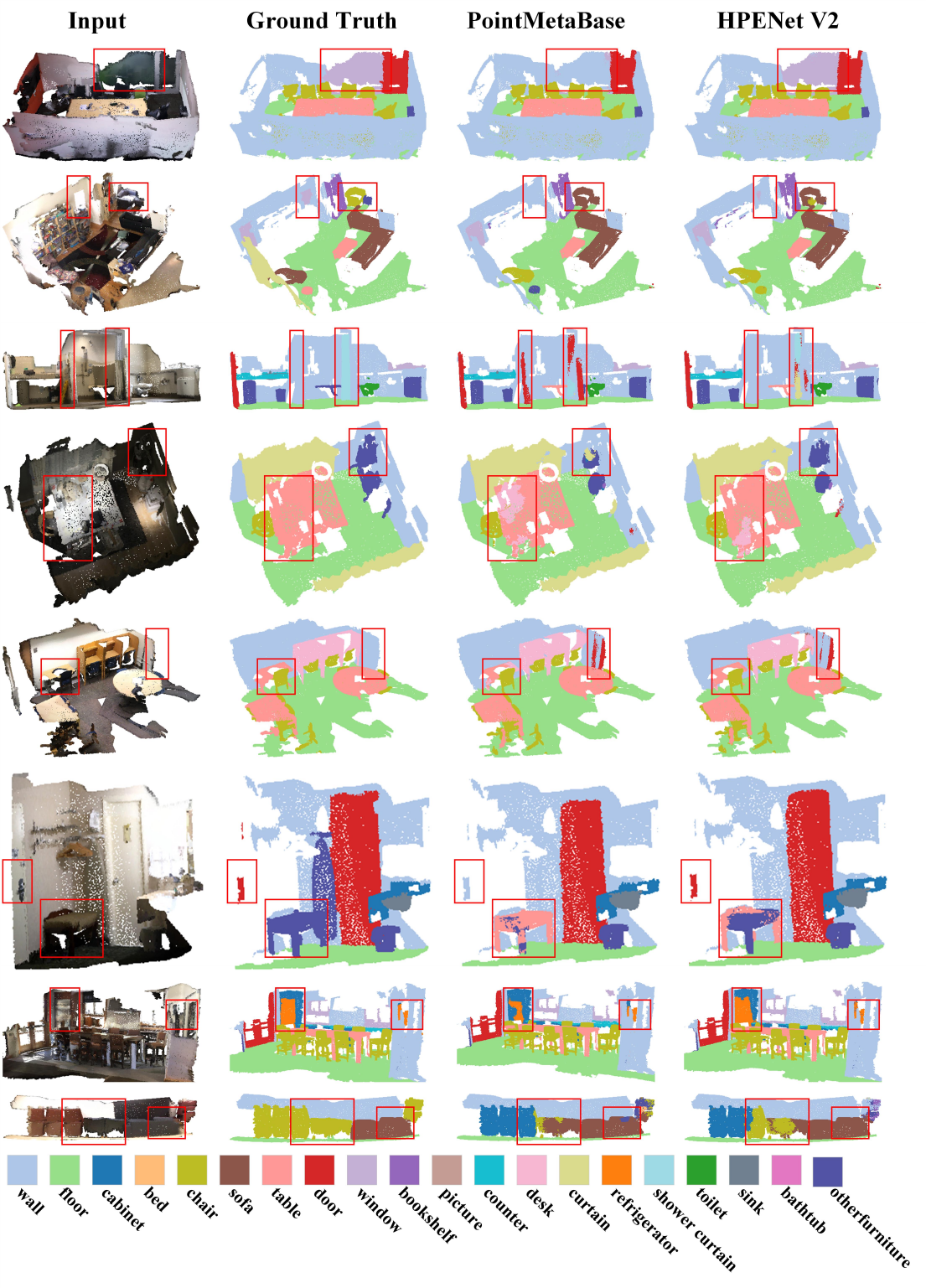}
	\end{center}
	\caption{Representative qualitative results of  HPENet V2 and the balanced MLP-based method PointMetaBase~\cite{98lin2023meta} on ScanNet V2. The main differences are highlighted in red frames.
	}
	\label{fig:scannetv1}
\end{figure*}

\vspace{0.7mm}
\noindent\textit{\textbf{More qualitative results on ScanNet V2.}} 
Fig.~\ref{fig:scannetv1} shows more qualitative results produced by HPENet V2 on ScanNet V2. As shown in Fig.~\ref{fig:scannetv1}, HPENet V2 makes predictions closer to the ground truth compared to the balanced MLP-based method PointMetaBase~\cite{98lin2023meta}. Specifically, HPENet V2 is able to segment wall ($2^{nd}$, $3^{rd}$, and $5^{th}$ rows), chair ($2^{nd}$ and $8^{th}$ rows), and table ($4^{th}$, and $5^{th}$ rows), while PointMetaBase fails to segment these objects appropriately in some instances.

%

\end{document}